\documentclass[10pt]{article} 
\usepackage[accepted]{tmlr}

\usepackage{xcolor}
\usepackage{amsmath,amsfonts,bm}
\definecolor{myblue}{rgb}{0.00, 0.45, 0.70}
\definecolor{mygreen}{rgb}{0.01, 0.62, 0.45}
\definecolor{myyellow}{rgb}{0.9, 0.5, 0}
\definecolor{mygrey}{rgb}{0.5, 0.5, 0.5}
\definecolor{myred}{rgb}{0.84, 0.37, 0.00}

\def\1{\bm{1}}

\newcommand{\RNum}[1]{\uppercase\expandafter{\romannumeral #1\relax}}

\def\vx{{\bm{x}}}
\def\vy{{\bm{y}}}

\DeclareMathAlphabet{\mathsfit}{\encodingdefault}{\sfdefault}{m}{sl}
\SetMathAlphabet{\mathsfit}{bold}{\encodingdefault}{\sfdefault}{bx}{n}

\usepackage[utf8]{inputenc} 
\usepackage[T1]{fontenc}    
\usepackage{hyperref}       
\usepackage{url}            
\usepackage{booktabs}       
\usepackage{amsfonts}       
\usepackage{nicefrac}       
\usepackage{xcolor}         
\usepackage{wrapfig}
\usepackage{graphicx}
\usepackage{caption}
\usepackage{subcaption}
\usepackage{booktabs}
\usepackage{tikz}
\usetikzlibrary{shapes.multipart}
\usetikzlibrary{calc}
\usetikzlibrary{backgrounds}

\usepackage{latexsym}
\usepackage{tabularx}
\usepackage{multirow}
\usepackage{enumitem}
\usepackage{amsmath}
\usepackage{pifont}
\usepackage{amssymb}
\usepackage{color, colortbl}
\usepackage{float}
\usepackage{dblfloatfix}
\usepackage{appendix}
\usepackage{bbold}
\usepackage{xspace}
\usepackage{adjustbox}
\usepackage{changepage}
\definecolor{DarkPink}{rgb}{0.5,0.0,0.18}
\definecolor{DarkGreen}{rgb}{0.1,0.5,0.1}
\definecolor{DarkRed}{rgb}{0.5,0.1,0.1}
\definecolor{DarkBlue}{rgb}{0.1,0.1,0.7}
\definecolor{DarkYellow}{rgb}{.79,.79,0}
\hypersetup{
    unicode=false,          
    pdftoolbar=true,        
    pdfmenubar=true,        
    pdffitwindow=false,      
    pdfnewwindow=true,      
    colorlinks=true,       
    linkcolor=DarkBlue,          
    citecolor=DarkPink,        
    filecolor=DarkRed,      
    urlcolor=DarkBlue,          
    pdftitle={},
    pdfauthor={},
}

\usepackage{amsmath}
\usepackage{amssymb}
\usepackage{mathtools}
\usepackage{amsthm}

\newcommand{\valstd}[2]{#1 {\tiny#2}}

\title{Leveraging the True Depth of LLMs}

\author{\name Ram\'on Calvo Gonz\'alez \email ramon.calvogonzalez@unige.ch \\
      \addr University of Geneva
      \AND
      \name Daniele Paliotta \\ 
      \addr University of Geneva
      \AND
      \name Matt\'eo Pagliardini \\
      \addr EPFL \\
      \AND
      \name Martin Jaggi \\
      \addr EPFL \\
      \AND
      \name Fran\c cois Fleuret \\
      \addr University of Geneva \\
      FAIR at META}

\def\month{01}  
\def\year{2026} 
\def\openreview{\url{https://openreview.net/forum?id=JccJ6YfWd4}} 

\makeatletter
\def\mathcolor#1#{\@mathcolor{#1}}
\def\@mathcolor#1#2#3{%
  \protect\leavevmode
  \begingroup
    \color#1{#2}#3%
  \endgroup
}
\makeatother

\definecolor{feature1}{RGB}{230,230,250} 
\definecolor{feature2}{RGB}{176,224,230} 
\definecolor{feature3}{RGB}{144,238,144} 
\definecolor{feature4}{RGB}{200,200,200}

\newcommand{\stackedrectangles}[5][(0,0)]{%
    
    \def\xshift{-0.2};
    \def\yshift{-0.2};
    
    \coordinate (base) at (#1);
    
    \coordinate (#2-A1) at (base);
    \coordinate (#2-B1) at ($(#2-A1)+(2,0)$);
    \coordinate (#2-C1) at ($(#2-B1)+(0,1.5)$);
    \coordinate (#2-D1) at ($(#2-A1)+(0,1.5)$);
    \coordinate (#2-E1bot) at ($(#2-A1)!0.5!(#2-B1)$);
    \coordinate (#2-E1right) at ($(#2-B1)!0.5!(#2-C1)$);
    \coordinate (#2-E1top) at ($(#2-C1)!0.5!(#2-D1)$);
    \coordinate (#2-E1left) at ($(#2-D1)!0.5!(#2-A1)$);
    \fill[#3] (#2-A1) rectangle (#2-C1);
    \draw[thick] (#2-A1) rectangle (#2-C1);
    
    \coordinate (#2-A2) at ($(#2-A1)+(\xshift,\yshift)$);
    \coordinate (#2-B2) at ($(#2-A2)+(2,0)$);
    \coordinate (#2-C2) at ($(#2-B2)+(0,1.5)$);
    \coordinate (#2-D2) at ($(#2-A2)+(0,1.5)$);
    \coordinate (#2-E2bot) at ($(#2-A2)!0.5!(#2-B2)$);
    \coordinate (#2-E2right) at ($(#2-B2)!0.5!(#2-C2)$);
    \coordinate (#2-E2top) at ($(#2-C2)!0.5!(#2-D2)$);
    \coordinate (#2-E2left) at ($(#2-D2)!0.5!(#2-A2)$);
    \fill[#3] (#2-A2) rectangle (#2-C2);
    \draw[thick] (#2-A2) rectangle (#2-C2);
    
    \coordinate (#2-A3) at ($(#2-A2)+(\xshift,\yshift)$);
    \coordinate (#2-B3) at ($(#2-A3)+(2,0)$);
    \coordinate (#2-C3) at ($(#2-B3)+(0,1.5)$);
    \coordinate (#2-D3) at ($(#2-A3)+(0,1.5)$);
    \coordinate (#2-E3bot) at ($(#2-A3)!0.5!(#2-B3)$);
    \coordinate (#2-E3right) at ($(#2-B3)!0.5!(#2-C3)$);
    \coordinate (#2-E3top) at ($(#2-C3)!0.5!(#2-D3)$);
    \coordinate (#2-E3left) at ($(#2-D3)!0.5!(#2-A3)$);
    \fill[#4] (#2-A3) rectangle (#2-C3);
    \draw[thick] (#2-A3) rectangle (#2-C3);
    
    \coordinate (#2-A4) at ($(#2-A3)+(\xshift,\yshift)$);
    \coordinate (#2-B4) at ($(#2-A4)+(2,0)$);
    \coordinate (#2-C4) at ($(#2-B4)+(0,1.5)$);
    \coordinate (#2-D4) at ($(#2-A4)+(0,1.5)$);
    \coordinate (#2-E4bot) at ($(#2-A4)!0.5!(#2-B4)$);
    \coordinate (#2-E4right) at ($(#2-B4)!0.5!(#2-C4)$);
    \coordinate (#2-E4top) at ($(#2-C4)!0.5!(#2-D4)$);
    \coordinate (#2-E4left) at ($(#2-D4)!0.5!(#2-A4)$);
    \fill[#4] (#2-A4) rectangle (#2-C4);
    \draw[thick] (#2-A4) rectangle (#2-C4);%

    \node[align=center] at ($(#2-A4)!0.5!(#2-C4)$) {#5};
}

\newcommand{\stackedroundedrectangles}[5][(0,0)]{%
    
    \def\xshift{-0.2}%
    \def\yshift{-0.2}%
    
    \coordinate (base) at (#1);
    
    \coordinate (#2-A1) at (base);
    \coordinate (#2-B1) at ($(#2-A1)+(2,0)$);
    \coordinate (#2-C1) at ($(#2-B1)+(0,1.5)$);
    \coordinate (#2-D1) at ($(#2-A1)+(0,1.5)$);
    \coordinate (#2-E1bot) at ($(#2-A1)!0.5!(#2-B1)$);
    \coordinate (#2-E1right) at ($(#2-B1)!0.5!(#2-C1)$);
    \coordinate (#2-E1top) at ($(#2-C1)!0.5!(#2-D1)$);
    \coordinate (#2-E1left) at ($(#2-D1)!0.5!(#2-A1)$);
    \fill[#3, rounded corners=5pt] (#2-A1) rectangle (#2-C1);
    \draw[thick, rounded corners=5pt] (#2-A1) rectangle (#2-C1);
    
    \coordinate (#2-A2) at ($(#2-A1)+(\xshift,\yshift)$);
    \coordinate (#2-B2) at ($(#2-A2)+(2,0)$);
    \coordinate (#2-C2) at ($(#2-B2)+(0,1.5)$);
    \coordinate (#2-D2) at ($(#2-A2)+(0,1.5)$);
    \coordinate (#2-E2bot) at ($(#2-A2)!0.5!(#2-B2)$);
    \coordinate (#2-E2right) at ($(#2-B2)!0.5!(#2-C2)$);
    \coordinate (#2-E2top) at ($(#2-C2)!0.5!(#2-D2)$);
    \coordinate (#2-E2left) at ($(#2-D2)!0.5!(#2-A2)$);
    \fill[#3, rounded corners=5pt] (#2-A2) rectangle (#2-C2);
    \draw[thick, rounded corners=5pt] (#2-A2) rectangle (#2-C2);
    
    \coordinate (#2-A3) at ($(#2-A2)+(\xshift,\yshift)$);
    \coordinate (#2-B3) at ($(#2-A3)+(2,0)$);
    \coordinate (#2-C3) at ($(#2-B3)+(0,1.5)$);
    \coordinate (#2-D3) at ($(#2-A3)+(0,1.5)$);
    \coordinate (#2-E3bot) at ($(#2-A3)!0.5!(#2-B3)$);
    \coordinate (#2-E3right) at ($(#2-B3)!0.5!(#2-C3)$);
    \coordinate (#2-E3top) at ($(#2-C3)!0.5!(#2-D3)$);
    \coordinate (#2-E3left) at ($(#2-D3)!0.5!(#2-A3)$);
    \fill[#4, rounded corners=5pt] (#2-A3) rectangle (#2-C3);
    \draw[thick, rounded corners=5pt] (#2-A3) rectangle (#2-C3);
    
    \coordinate (#2-A4) at ($(#2-A3)+(\xshift,\yshift)$);
    \coordinate (#2-B4) at ($(#2-A4)+(2,0)$);
    \coordinate (#2-C4) at ($(#2-B4)+(0,1.5)$);
    \coordinate (#2-D4) at ($(#2-A4)+(0,1.5)$);
    \coordinate (#2-E4bot) at ($(#2-A4)!0.5!(#2-B4)$);
    \coordinate (#2-E4right) at ($(#2-B4)!0.5!(#2-C4)$);
    \coordinate (#2-E4top) at ($(#2-C4)!0.5!(#2-D4)$);
    \coordinate (#2-E4left) at ($(#2-D4)!0.5!(#2-A4)$);
    \fill[#4, rounded corners=5pt] (#2-A4) rectangle (#2-C4);
    \draw[thick, rounded corners=5pt] (#2-A4) rectangle (#2-C4);
    \node[align=center] at ($(#2-A4)!0.5!(#2-C4)$) {#5};
}

\newcommand{\stackedtworectangles}[5][(0,0)]{%
    
    \def\xshift{-0.2};
    \def\yshift{-0.2};
    
    \coordinate (base) at (#1);
    
    \coordinate (#2-A1) at (base);
    \coordinate (#2-B1) at ($(#2-A1)+(2,0)$);
    \coordinate (#2-C1) at ($(#2-B1)+(0,1.5)$);
    \coordinate (#2-D1) at ($(#2-A1)+(0,1.5)$);
    \coordinate (#2-E1bot) at ($(#2-A1)!0.5!(#2-B1)$);
    \coordinate (#2-E1right) at ($(#2-B1)!0.5!(#2-C1)$);
    \coordinate (#2-E1top) at ($(#2-C1)!0.5!(#2-D1)$);
    \coordinate (#2-E1left) at ($(#2-D1)!0.5!(#2-A1)$);
    \fill[#3] (#2-A1) rectangle (#2-C1);
    \draw[thick] (#2-A1) rectangle (#2-C1);
    
    \coordinate (#2-A2) at ($(#2-A1)+(\xshift,\yshift)$);
    \coordinate (#2-B2) at ($(#2-A2)+(2,0)$);
    \coordinate (#2-C2) at ($(#2-B2)+(0,1.5)$);
    \coordinate (#2-D2) at ($(#2-A2)+(0,1.5)$);
    \coordinate (#2-E2bot) at ($(#2-A2)!0.5!(#2-B2)$);
    \coordinate (#2-E2right) at ($(#2-B2)!0.5!(#2-C2)$);
    \coordinate (#2-E2top) at ($(#2-C2)!0.5!(#2-D2)$);
    \coordinate (#2-E2left) at ($(#2-D2)!0.5!(#2-A2)$);
    \fill[#4] (#2-A2) rectangle (#2-C2);
    \draw[thick] (#2-A2) rectangle (#2-C2);

    \node[align=center] at ($(#2-A2)!0.5!(#2-C2)$) {#5};
}

\newcommand{\stackedroundedtworectangles}[5][(0,0)]{%
    
    \def\xshift{-0.2}%
    \def\yshift{-0.2}%
    
    \coordinate (base) at (#1);
    
    \coordinate (#2-A1) at (base);
    \coordinate (#2-B1) at ($(#2-A1)+(2,0)$);
    \coordinate (#2-C1) at ($(#2-B1)+(0,2)$);
    \coordinate (#2-D1) at ($(#2-A1)+(0,2)$);
    \coordinate (#2-E1bot) at ($(#2-A1)!0.5!(#2-B1)$);
    \coordinate (#2-E1right) at ($(#2-B1)!0.5!(#2-C1)$);
    \coordinate (#2-E1top) at ($(#2-C1)!0.5!(#2-D1)$);
    \coordinate (#2-E1left) at ($(#2-D1)!0.5!(#2-A1)$);
    \fill[#3, rounded corners=5pt] (#2-A1) rectangle (#2-C1);
    \draw[thick, rounded corners=5pt] (#2-A1) rectangle (#2-C1);
    
    \coordinate (#2-A2) at ($(#2-A1)+(\xshift,\yshift)$);
    \coordinate (#2-B2) at ($(#2-A2)+(2,0)$);
    \coordinate (#2-C2) at ($(#2-B2)+(0,2)$);
    \coordinate (#2-D2) at ($(#2-A2)+(0,2)$);
    \coordinate (#2-E2bot) at ($(#2-A2)!0.5!(#2-B2)$);
    \coordinate (#2-E2right) at ($(#2-B2)!0.5!(#2-C2)$);
    \coordinate (#2-E2top) at ($(#2-C2)!0.5!(#2-D2)$);
    \coordinate (#2-E2left) at ($(#2-D2)!0.5!(#2-A2)$);
    \fill[#4, rounded corners=5pt] (#2-A2) rectangle (#2-C2);
    \draw[thick, rounded corners=5pt] (#2-A2) rectangle (#2-C2);

    \node[align=center] at ($(#2-A2)!0.5!(#2-C2)$) {#5};
}

\begin{document}

\maketitle

\begin{abstract}
The remarkable capabilities of Large Language Models (LLMs) are overshadowed by their immense
computational cost. While recent work has shown that many LLM layers can be reordered or even
removed with minimal impact on accuracy, these insights have not been translated into significant
inference speedups. To bridge this gap, we introduce a novel method that restructures the
computational graph by grouping and evaluating consecutive layer pairs in parallel. This approach,
requiring no retraining, yields a 1.19x throughput gain on Llama 2 7B while reducing the average
benchmark accuracy by only 1.5\%. We demonstrate the practical value of this method for large-scale
LLM deployment and show that some of the lost accuracy can be recovered with lightweight
fine-tuning of the parallelized layers.
\end{abstract}

\section{Introduction}

\begin{wrapfigure}[14]{r}{0.45\textwidth}  
  \vspace{-2\baselineskip}
  \centering
  \begin{tikzpicture}
    \node {\includegraphics[width=0.9\linewidth,trim=20 20 20 0]{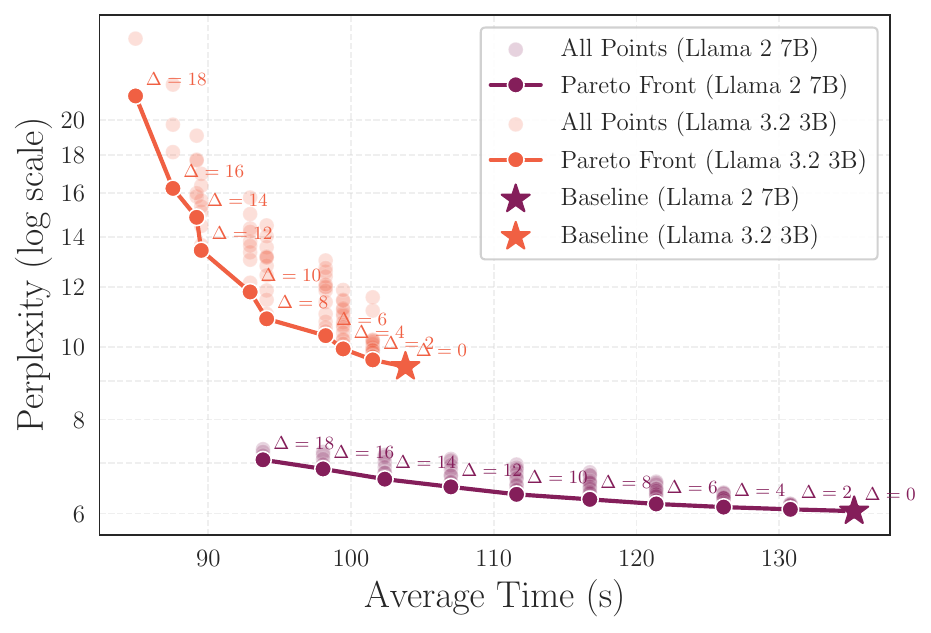}};
    \draw[<->,thick] (-1.1,-1.0)--++(0,0.5);
  \end{tikzpicture}
  \caption{\label{fig:pareto}
    The effect of LP on execution time (4K tokens) and perplexity (measured against RedPajama \citep{together2023redpajama}). 
    }
\end{wrapfigure}

The rapid advancement of LLMs has revolutionized Artificial Intelligence
applications across industries. However, the ever-increasing computational
demands of these models, with parameters often numbering hundreds of
billions, present significant commercial challenges. Efficient inference
is crucial for organizations that deploy these models at scale, as it
directly impacts operational costs, user experience, and environmental
sustainability \citep{singh2025surveysustainabilitylargelanguage,
xu2024surveyresourceefficientllmmultimodal,
wu2022sustainableaienvironmentalimplications}. Monthly cloud computing expenses
for LLM inference can reach millions of dollars for high-traffic applications,
making optimization techniques essential. In addition, reducing inference
latency is critical for real-time applications and for deploying models on
devices with more limited compute.

\begin{wrapfigure}[24]{R}{0.45\textwidth}
\centering
\begin{tikzpicture}[
    scale=0.85,
    block/.style={rectangle, draw, minimum width=1.5cm, minimum height=1cm},
    plus/.style={circle, draw, minimum size=0.6cm},
    arrow/.style={->, thick},
    skip/.style={->, thick}
]
\node[plus] (plus2) at (0,6.6) {+};
\node[block] (mlp) at (0,5.5) {FFN};
\node[block] (ln2) at (0,4) {LN};
\node[block] (mha) at (0,1.5) {MHA};
\node[block] (ln1) at (0,0) {LN};
\node[plus] (plus1) at (0,2.6) {+};
\draw[arrow] (0,-1) -- (ln1);  
\draw[arrow] (ln1) -- (mha);
\draw[arrow] (mha) -- (plus1);
\draw[arrow] (plus1) -- (ln2);
\draw[arrow] (ln2) -- (mlp);
\draw[arrow] (mlp) -- (plus2);
\draw[skip] (0,-.75) -| (1.5,2.6) -- (plus1);
\draw[skip] (0, 3) -| (1.5,6.6) -- (plus2);
\draw[skip] (plus2) -- (0, 7.2);

\draw[thick] (2.15,-1.25) -- (2.15,7);

\node[plus] (plus2_var) at (4.4,6.75) {+};
\node[block] (mlp1) at (3.4,5.5) {FFN1};
\node[block] (ln_mlp1) at (3.4,4) {LN};
\node[block] (mlp2) at (5.4,5.5) {FFN2};
\node[block] (ln_mlp2) at (5.4,4) {LN};
\node[block] (mha1) at (3.4,1.5) {MHA1};
\node[block] (ln_mha1) at (3.4,0) {LN};
\node[block] (mha2) at (5.4,1.5) {MHA2};
\node[block] (ln_mha2) at (5.4,0) {LN};
\node[plus] (plus_mha) at (4.4,2.75) {+};
\draw[arrow] (4.4,-1) -| (ln_mha1);
\draw[arrow] (4.4,-1) -| (ln_mha2);
\draw[arrow] (ln_mha1) -- (mha1);
\draw[arrow] (ln_mha2) -- (mha2);
\draw[arrow] (mha1) --  (plus_mha);
\draw[arrow] (mha2) --  (plus_mha);
\draw[arrow] (ln_mlp1) -- (mlp1);
\draw[arrow] (ln_mlp2) -- (mlp2);
\draw[arrow] (mlp1) -- (plus2_var);
\draw[arrow] (mlp2) -- (plus2_var);
\draw[arrow] (plus_mha) -- (4.4,3.25) -| (ln_mlp1);
\draw[arrow] (plus_mha) -- (4.4,3.25) -| (ln_mlp2);
\draw[arrow] (plus2_var) -- (4.4,7.25);
\draw[skip] (4.4,-1) -| (6.4,2.75) -- (plus_mha);
\draw[skip] (4.4,3.25) -| (6.4,6.75) -- (plus2_var);
\draw[thick] (4.4,-1.5) -- (4.4,-1);

\node (a) at (0, -1.75) {(a)};
\node (b) at (4.4, -1.75) {(b)};
\end{tikzpicture}
\caption{\label{fig:normal_and_pl_transformer_layer} Comparison of a normal transformer block (a) with our layer parallel implementation (b). Divergent paths in (b) are split across the Tensor Parallel axis (Eq.~\ref{eq:lp}).}
\end{wrapfigure}

Thus, the development and implementation of efficient inference methods
has become a key differentiator in the competitive AI market, driving both
innovation and profitability.

LLMs have evolved to incorporate architectures with hundreds of
layers \citep{gpt4, grattafiori2024llama3herdmodels}. These models are
constructed from stacked transformer blocks, each comprising attention
and feedforward subblocks, with a residual stream traversing the entire
architecture to facilitate efficient gradient propagation during training.
This architectural choice parallels the design principles of ResNets
\citep{he2015deepresiduallearningimage}, where research has shown that
network depth may be partially redundant, allowing layer reordering
without significant performance loss \citep{VeitWB16}. Recent
investigations have revealed similar flexibility in transformer architectures
\citep{lad2024remarkablerobustnessllmsstages}, where interventions such as
layer removal and swapping are applied without large performance degradations.
Although these findings challenge our understanding of LLMs' true effective
depth, their potential for optimizing inference efficiency remains unexplored.

Inspired by this observed layer independence, we investigated several
interventions to the computational graph of pre-trained LLMs. Our exploration of
layer shuffling, pruning, and merging revealed that multiple consecutive block pairs
can be processed in parallel while maintaining accuracy across perplexity and In-Context
Learning (ICL) benchmarks. This led us to propose Layer Parallelism (LP), a novel approach
that enhances inference speed when performing inference in the Tensor Parallel (TP) regime.
LP modifies the computational graph of a pre-trained LLM to reduce the inter-device
communication by half, with a minimal drop in model performance. Furthermore, we show that
this performance degradation can be partially mitigated through targeted fine-tuning procedures.

\textbf{Contributions.} Our contributions can be summarized as follows:
\begin{itemize}
    \item We explore the space of interventions on the layers of a pre-trained LLM and find that some transformations, such as contiguous parallelization, preserve model performance.
    \item We find that we can define a parallelization transform on the computational graph of two sequential Transformer layers, and stack this parallelization operation across several sequential pairs of layers without losing significant ICL performance. Our approach, which we call LP, can be applied to existing Transformer models.
    \item We show that by fine-tuning the LP blocks we can recover some of the lost performance, while retaining the previously obtained speed-up.
\end{itemize}




\section{Related work}

\textbf{The effective depth of Deep Networks.} Theoretically, given enough
width, any feed-forward network with at least one hidden layer can model
any function \citep{Pinkus_1999}. In practice, it is easier to achieve
high expressivity by increasing the model's depth. However, naively
increasing network depth can complicate optimization, since the gradients
now have to flow through many layers. To alleviate this problem, ResNets
\citep{he2015deepresiduallearningimage} introduced skip connections at
regular intervals to allow an easy flow of the gradient to the first layers.
Alternatively, Inception \citep{szegedy2014goingdeeperconvolutions} explored
approaches to boost computational power by adding additional processing units
along different parallel pathways in the computational network, rather than
just along a single sequential path. A unification of both methods can be
found in the Highway Networks \citep{srivastava2015highwaynetworks}, where the
skip connection of the residual blocks consists of another block of compute.
Nowadays, residual connections are ubiquitous in large models.

\textbf{Efficient inference of LLMs.} Several complementary approaches
exist for enhancing the computational efficiency of large-scale models,
primarily through pruning/sparsity, quantization, and parallelism.
Pruning \citep{LeCun1989OptimalBD, han2015learningweightsconnectionsefficient,
han2016deepcompressioncompressingdeep, frantar2023sparsegpt} constitutes a
dimensional reduction methodology that systematically eliminates redundant
parameters while preserving model performance, thereby introducing
architectural sparsity. This methodology is founded on empirical evidence
demonstrating that neural networks frequently exhibit overparameterization,
containing numerous weights with negligible contributions to the
output. Through sophisticated pruning strategies, the inherent sparsity
support in contemporary accelerators can be leveraged to enhance both
memory utilization and computational efficiency \citep{zhang2020sparch,
Spatten21}. Early-exit \citep{teerapittayanon2017branchynetfastinferenceearly,
zhou2020bertlosespatiencefast} can be seen as a way of runtime layer-wise
pruning, which halts the LLM forward pass when the next token certainty
is high in the intermediate layers. This approach can also be thought
of as a way of reducing the effective depth of the model at test time. In
contrast, quantization encompasses the transformation of floating-point
numerical representations (predominantly FP32) into reduced-precision
integer formats, such as INT8 or INT4 \citep{shen2019qberthessianbasedultra,
han2016deepcompressioncompressingdeep, jacob2018quantization}. When implemented
on hardware accelerators, these lower-precision representations allow for higher
FLOPs and better use of the memory bandwidth, addressing a primary bottleneck in
modern large-scale models \citep{gholami2024ai}; moreover, integer-based
computations yield enhanced processing speed and substantially improved energy
efficiency \citep{horowitz20141}. Finally, parallelization techniques during
inference, such as tensor and pipeline parallelism, enable the distribution of
computational workload across multiple accelerators, thereby reducing latency
and increasing throughput, although this often requires careful consideration of
communication overhead and load balancing \citep{li2024tpillmserving70bscalellms,
narayanan2021efficientlargescalelanguagemodel}.

\textbf{Tensor-parallel optimizations.} Inter-device communication remains the dominant bottleneck in tensor parallelism, as each transformer sub-module imposes a synchronization step. Recent work targets this cost by either removing synchronization points or shrinking the communicated tensors. Sync-Point Drop \citep{kim2025spdsyncpointdropefficient} cuts the number of all-reduce operations by modifying block structure so that local attention outputs propagate without immediate aggregation; layers are ranked by sensitivity to synchronization removal, and only the sensitive ones receive targeted tuning, yielding a 20\% speedup with roughly 1\% accuracy loss. A complementary direction reduces communication volume. \citep{dong2024lowbitcommunicationtensorparallel} quantizes per-layer activation exchanges, reaching a 3.8× compression ratio with about a 2\% drop on the evaluated benchmarks.

\textbf{Parallel attention-feedforward fusion.} GPT-J \citep{gptj} introduced a
parallel formulation of the transformer decoder layer, executing attention and
feedforward sub-blocks concurrently:
\begin{align*}
y &= x + \operatorname{MHA}(\operatorname{LN}_{\mathrm{MHA}}(x)) + \operatorname{FFN}(\operatorname{LN}_{\mathrm{FFN}}(x))
\end{align*}
For models trained with tensor parallelism, this modification halves the
number of required all-reduce operations. Additionally, it reduces memory
bandwidth usage by eliminating one read and write operation of hidden states
from HBM. The input projections for the attention and MLP sub-blocks can
also be fused into a single kernel, further increasing arithmetic density.
Consequently, training time is reduced by approximately 15\% without observable
performance degradation. PaLM \citep{chowdhery2023palm} also adopted this
formulation, noting that negative effects from deviating from the standard
transformer self-attention diminish with increasing model size. This parallel
formulation continues to be employed in more recent LLMs, including Gemini 1.5
Flash \citep{team2024gemini}.

In contrast to these methods, which require training a new model from scratch
with a modified architecture, our approach is applied post-hoc to already-trained
models \citep{touvron2023llama2, grattafiori2024llama3herdmodels,
yang2025qwen3technicalreport}. This highlights two other key differences. First, the
granularity of parallelism differs: GPT-J-style models parallelize the attention and
feed-forward sub-blocks \textit{within} a single layer, whereas our method parallelizes
\textit{entire consecutive layers}, directly reducing the model's effective depth. Second,
LP accepts a trade-off by approximating the original computation, which results in a slight
performance degradation in exchange for inference acceleration. This degradation can be
largely recovered with light fine-tuning. The GPT-J architecture, by contrast, is exact
by definition, as the model was trained with it from the beginning.


\textbf{Parallelism via Computational Graph Optimization.} Recent research
has investigated architectural layer-level optimization strategies to
enhance transformer model inference efficiency. The Staircase Transformer
\citep{cutler2025stagformertimestaggeringtransformer} implements parallel
layer execution with dynamic recurrent computation based on model requirements.
Similarly, the Staggering Transformer \citep{cai2024medusasimplellminference}
achieves layer parallelization by connecting layer $l_k$ at time step $t$
to both the $(t-1)$ output of layer $l_{k-1}$ and the $t$ output of layer
$l_{k-2}$. To the best of our knowledge, no research has addressed the fusion of
consecutive layers through tensor parallelism.

\section{Effective Depth}
\label{sec:effective-depth}

\begin{wrapfigure}{r}{0.5\textwidth}
    \centering
    \begin{tikzpicture}[scale=0.65,
        box/.style={draw, rectangle, minimum width=1.2cm, minimum height=0.75cm},
        merged/.style={draw, rectangle, minimum width=1.4cm, minimum height=0.75cm},
        arrow/.style={->, thick},
        darrow/.style={<->, thick, dashed},
    ]
    
    \node[box] (input_left) at (0,0) {Input};
    \node[box] (layer1_left) at (0,1.3) {L2};
    \node[box] (layer2_left) at (0,2.7) {L1};
    \node[box] (output_left) at (0,4.05) {Output};
    
    \draw[arrow] (input_left) -- (layer1_left);
    \draw[arrow] (layer1_left) -- (layer2_left);
    \draw[arrow] (layer2_left) -- (output_left);
    \draw[darrow] (layer1_left) -| +(1.25,1.25) -- (layer2_left);
    \node (label_a) at (0, -1) {(a)};
    
    
    \node[box] (input_middle) at (3,0) {Input};
    \node[merged] (layers_middle) at (3,2.025) {$\frac{1}{2}$L1 + $\frac{1}{2}$L2};
    \node[box] (output_middle) at (3,4.05) {Output};
    
    \draw[arrow] (input_middle) -- (layers_middle);
    \draw[arrow] (layers_middle) -- (output_middle);
    \node (label_d) at (3, -1) {(b)};
    
    
    \node[box] (input_right) at (6,0) {Input};
    \node[box] (layer1_right) at (6,1.35) {L1};
    \draw[red, thick] (5.5,0.95) -- (6.5,1.55);
    \draw[red, thick] (5.5,1.55) -- (6.5,0.95);
    \node[box] (layer2_right) at (6,2.7) {L2};
    \node[box] (output_right) at (6,4.05) {Output};
    
    \draw[arrow] (layer2_right) -- (output_right);
    \draw[arrow] (input_right) -- (7.25,0) -- (7.25,2.7) -- (layer2_right);
    \node (label_c) at (6, -1) {(c)};
    
    
    \node[box] (input_far_right) at (9,0) {Input};
    \node[merged] (layers_far_right) at (9,2.025) {};
    \draw[thick] (9,0.75) -- (9,2.58);
    \node at (8.5,1.875) {L1};
    \node at (9.5,1.875) {L2};
    \node[box] (output_far_right) at (9,4.05) {Output};
    
    \draw[arrow] (input_far_right) -- (layers_far_right);
    \draw[arrow] (layers_far_right) -- (output_far_right);
    \node (label_b) at (9, -1) {(d)};
    
    \end{tikzpicture}
    \caption{\textbf{Diagram of transformations applied in \S~\ref{sec:effective-depth}.} Diagrams \textbf{(a,b,c,d)} represent shuffling, merging, pruning and parallel respectively.}
    \label{fig:transform-diagrams}
\end{wrapfigure}

In this section we investigate the effective depth of pretrained LLMs by applying several
transformations and measuring the resulting perplexity degradation. We reveal loose dependencies
between intermediate layers. The transformations consist of shuffling, merging, and pruning
transformer layers. To avoid the combinatorial explosion resulting from considering all possible
subsets of transformer layers, we instead apply our transformations to all contiguous stretches of
layers. If $L=\{\ell_1, \ldots,\ell_N\}$ are the ordered layers, then we apply our transformations
to all the sublists $\{\ell_i\}_{i=s}^e$ with $1\leq s \leq e\leq N$. Previous works have shown
that---at least when considering pruning---the importance of layers is well-behaved, with low-importance layers close to one another \citep{men2024shortgptlayerslargelanguage}, which justifies
considering contiguous stretches of layers only.

\begin{figure*}[t]
    \centering
    \begin{subfigure}[b]{0.212\linewidth}
        \includegraphics[width=\textwidth,clip]{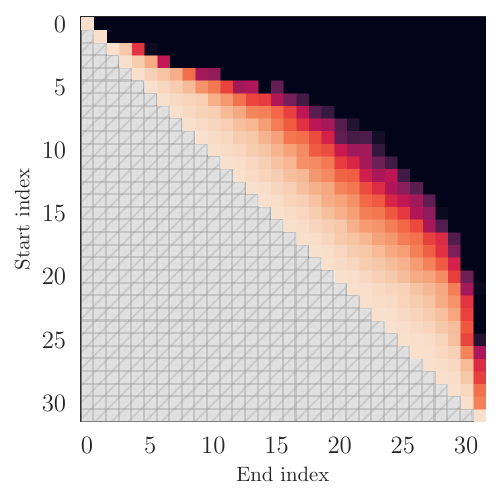}
        \caption{Shuffling.}
        \label{fig:matrix-shuffling}
    \end{subfigure}
    \begin{subfigure}[b]{0.18\linewidth}
        \includegraphics[width=\textwidth,trim={1.3cm 0cm 0cm 0cm},clip]{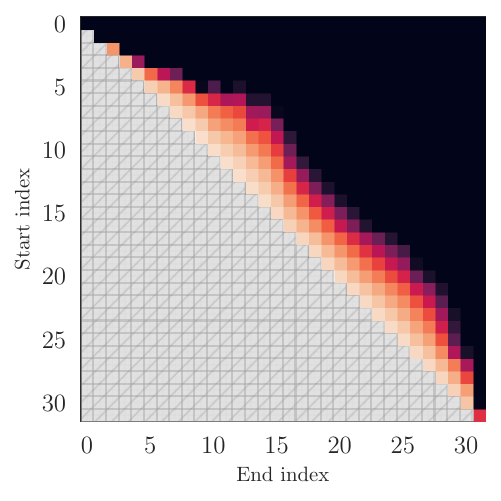}
        \caption{Pruning.}
        \label{fig:matrix-pruning}
    \end{subfigure}
    \begin{subfigure}[b]{0.18\linewidth}
        \includegraphics[width=\textwidth,trim={1.3cm 0cm 0cm 0cm},clip]{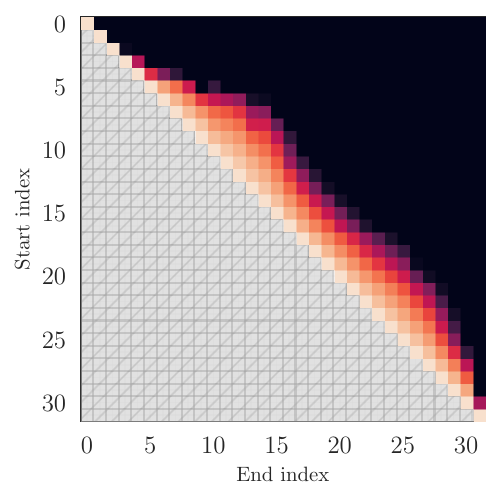}
        \caption{Merging.}
        \label{fig:matrix-merge}
    \end{subfigure}
    \begin{subfigure}[b]{0.18\linewidth}
        \includegraphics[width=\textwidth,trim={1.3cm 0cm 0cm 0cm},clip]{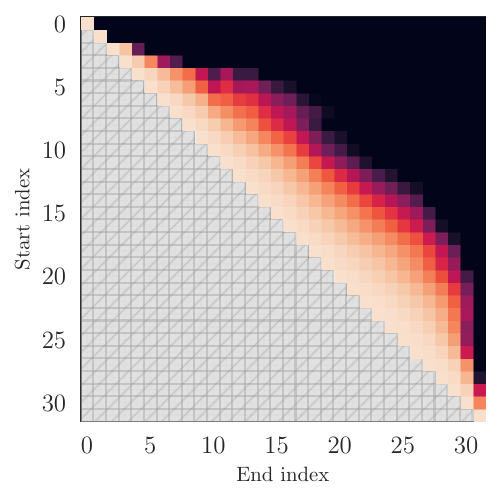}
        \caption{Parallel.}
        \label{fig:matrix-parallel}
    \end{subfigure}
    \begin{subfigure}[b]{0.215\linewidth}
        \includegraphics[width=\textwidth,trim={1.3cm 0cm 0cm 0cm},clip]{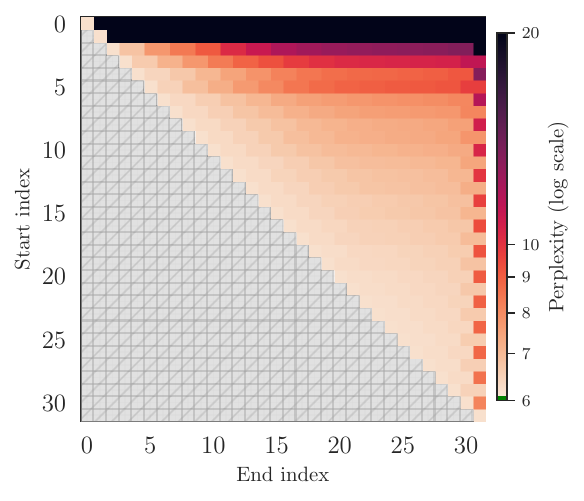}
        \caption{2-Parallel.}
        \label{fig:matrix-2parallel}
    \end{subfigure}
    \caption{\textbf{Changes in perplexity when applying transformations on contiguous stretches of layers.} Each of the five heatmaps above corresponds to a transformation of a group of consecutive layers, where the row index $s$ corresponds to the first layer of the group, and the column index $e$ to the last. The color coding indicates how the perplexity---estimated on a subset of RedPajama \citep{together2023redpajama}---is impacted by the corresponding modification of the model. The perplexity for the base Llama 2 7B model is $6.2$. In \textbf{(a)}, we shuffle---for each forward---the layers from $s$ to $e$. We can see that many consecutive layers can be shuffled with little impact on the overall perplexity. For instance, shuffling layers $15$ to $25$---$10$ layers in total---raises the perplexity only to $9.1$. In \textbf{(b)}, we prune contiguous stretches of layers. We can see that not many blocks can be removed without starting to significantly degrade the perplexity. In \textbf{(c)} we merge contiguous layers. The results with merging are nearly identical to those for pruning. This reveals there is no advantage in merging layers, most likely a result of averaging matrices that originate from different initial values. In \textbf{(d)} we run contiguous blocks in parallel. Given the success of shuffling, it makes sense that this approach works well. Running blocks $17$ to $27$ raises the perplexity to $9.3$. Finally, in \textbf{(e)} we run \emph{pairs of consecutive layers} in parallel. As a result, we can parallelize much longer stretches of layers. For instance, we can apply this transformation from layer $4$ to $29$ and only increase the perplexity to $9.1$. This reduces the depth of the model from $32$ to $19$. This result makes it possible for us to leverage this parallelism for faster inference as we discuss in \S~\ref{sec:efficiency}.      
    }
    \label{fig:matrices}
\end{figure*}

\textbf{Shuffling, pruning and merging blocks.} We start by investigating
the effect of several transformations on the model's perplexity.
First, we experiment with shuffling contiguous stretches of layers
(Fig.~\ref{fig:transform-diagrams}a), re-ordering them according to random
permutations. Results, shown in Fig.~\ref{fig:matrices}(a), reveal that while
shuffling the early and late layers is detrimental, there are large stretches
of intermediate blocks that can be shuffled with surprisingly low impact on
perplexity. For instance, one can shuffle layers 15 through 24 of Llama 2
7B and only increase perplexity by 2.9. This suggests that many layers may
operate at a similar level of abstraction, challenging the classical belief of
strictly hierarchical representations. This observed layer decoupling is a key
insight. We also experiment with pruning (Fig.~\ref{fig:transform-diagrams}c)
and merging (Fig.~\ref{fig:transform-diagrams}b) contiguous layers. Pruning,
studied in prior works \citep{gromov2024unreasonableineffectivenessdeeperlayers,
jung2019compactassessingcompactnessrepresentations}, involves removing
layers entirely. Merging involves averaging the weights of consecutive
layers. We find that both transformations lead to a more significant
perplexity increase compared to shuffling (see Fig.~\ref{fig:matrices}b and
Fig.~\ref{fig:matrices}c). Merging, in particular, offers no advantage over
pruning, suggesting that naively combining weights from different layers is
ineffective. These initial experiments indicate that while layers are robust to
reordering, their individual parameters are crucial.

\textbf{Running blocks in parallel.} The observed layer decoupling suggests
that specific transformer operations may be executed independently, providing
an opportunity for parallel computation. More precisely, let's consider two
sequential transformer layers $\ell_{k}$ and $\ell_{k+1}$, each comprising
attention and Feed-Forward Network (FFN) sub-blocks ($A_k(\cdot)$ and
$F_k(\cdot)$, respectively). The standard sequential output $\vy$ for these
layers, given an input $\vx$, is given by:
\begin{align}
    \vy = \vx &+ \text{A}_k(\vx) \notag \\ 
              &+ \text{F}_k(\vx+ \text{A}_k(\vx)) \notag \\
              &+ \text{A}_{k+1}(\vx + \mathcolor{red}{\text{A}_k(\vx)} + \mathcolor{red}{\text{F}_k(\vx+ \text{A}_k(\vx))}) \notag \\
              &+ \text{F}_{k+1}(\vx + \mathcolor{red}{\text{A}_k(\vx)} + \mathcolor{red}{\text{F}_k(\vx+ \text{A}_k(\vx))} \\
              &+\text{A}_{k+1}(\vx + \mathcolor{red}{\text{A}_k(\vx)} + \mathcolor{red}{\text{F}_k(\vx+ \text{A}_k(\vx))})) \tag{SEQ}\label{eq:seq}
\end{align}
The highlighted terms represent the first block's contribution to the second block's processing. Given the observed layer independence, we can hypothesize that these terms have minimal impact, leading to the following approximation:
\begin{align}
    \hat{\vy} &= \vx + \text{A}_k(\vx) + \text{F}_k(\vx+ \text{A}_k(\vx)) + \text{A}_{k+1}(\vx) + \text{F}_{k+1}(\vx + \text{A}_{k+1}(\vx)) \tag{PAR}\label{eq:par} \\
    &\approx \vx + \text{A}_k(\vx) + \text{A}_{k+1}(\vx) + \text{F}_{k}(\vx + \text{A}_k(\vx) + \text{A}_{k+1}(\vx)) + \text{F}_{k+1}(\vx + \text{A}_k(\vx) + \text{A}_{k+1}(\vx)) \tag{LP}\label{eq:lp}
\end{align}
This approximation enables parallel execution of blocks $\ell_{k}$ and $\ell_{k+1}$ through divergent computational paths. 
We experiment with running contiguous stretches of layers in parallel and show our results in Fig.~\ref{fig:matrix-parallel}. We observe results similar to shuffling. Unlike shuffling, this approach allows us to potentially improve the runtime through enhanced parallelism. We show how we can, for instance, run layers $17$ to $27$ in parallel, only losing $3.1$ perplexity points, while reducing the depth of the model from $32$ to $23$. 

\begin{figure}[t]
\centering
\includegraphics[width=0.8\textwidth]{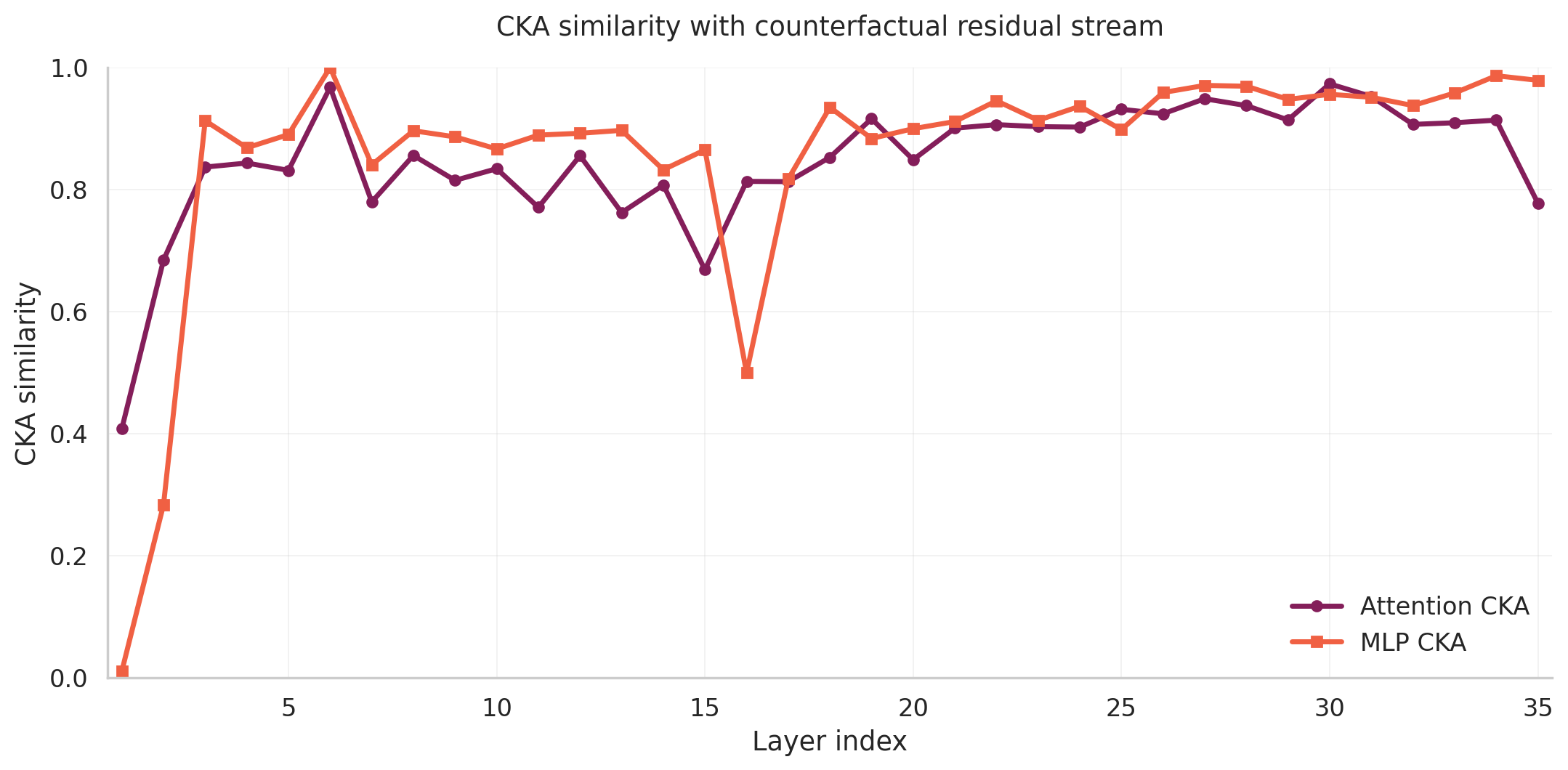}
\caption{\label{fig:cka}CKA similarity for Qwen3-4B between the original MHA/FFN activations and the counterfactual activations that exclude incoming residuals. Higher values imply greater invariance to the upstream residual stream. A plateau of high CKA similarity between pairs of layers is preceded by a sharp similarity decline at layer 16, which coincides with the performance degradations experienced when applying different levels of LP at different positions.}
\end{figure}

To assess how strongly attention and FFN sub-blocks rely on the residual stream generated by preceding layers, we apply a CKA-based comparison \citep{kornblith2019similarityneuralnetworkrepresentations}. For each prompt drawn from a small RedPajama subset, we record module activations under two conditions: a standard forward pass and a counterfactual pass in which the incoming residual contribution is removed before processing the next block. The standard decoder update for layer
$k$ is:
\begin{align*}
h_k &= x + A_k(x) + F_k\big(x + A_k(x)\big) \\
A_{k+1} &= A_{k+1}(h_k) \\
F_{k+1} &= F_{k+1}\big(h_k + A_{k+1}\big) \\
h_{k+1} &= h_k + A_{k+1} + F_{k+1} 
\end{align*}
Counterfactual activations are obtained by subtracting the residual stream before evaluating the next block:
\begin{align*}
\tilde{h}_k &= h_k - A_k(x) - F_k(x + A_k(x)) \\
\tilde{A}_{k+1} &= A_{k+1}(\tilde{h}_k) \\
\tilde{F}_{k+1} &= F_{k+1}(\tilde{h}_k + \tilde{A}_{k+1})
\end{align*}
Similarity between original and counterfactual activations is then computed via CKA:
\begin{align*}
\mathrm{S}_{A}^k &= \operatorname{CKA}(A_k, \tilde{A}_k) \\
\mathrm{S}_{F}^k &= \operatorname{CKA}(F_k, \tilde{F}_k)
\end{align*}
As shown in Fig.~\ref{fig:cka}, layers that tolerated the 2-parallel intervention also show high CKA similarity, indicating limited dependence of their attention and FFN computations on the immediate residual input.

\textbf{Contiguous 2-parallel.} Instead of parallelizing long stretches of layers, we experiment with running \emph{pairs of consecutive layers} in parallel. This springs from the assumption that local ordering matters less than global ordering, i.e. shuffling consecutive layers introduces fewer potential issues than shuffling layers separated by larger distances. As an example, if we apply the proposed transformation to layers $\{\ell_{15},\ell_{16},\ell_{17},\ell_{18},\ell_{19}\}$, it would result in the following process: (1) the two layers $\{\ell_{15},\ell_{16}\}$ process the input in parallel (according to equation \eqref{eq:par}), (2) the output is forwarded to layers $\{\ell_{17},\ell_{18}\}$ which process it in parallel; finally, in (3) their joint output is fed to layer $\ell_{19}$ which processes it on its own as any normal layer. The effect of such a transformation on the compute graph can be seen in Fig.~\ref{fig:matrix-2parallel}. Remarkably, it is possible to run wide stretches of consecutive pairs of blocks in parallel with only a minor degradation of perplexity. For instance, one can apply this transformation from layer $4$ to layer $29$ with only a perplexity increase of $2.9$, while reducing the model depth from $32$ to $19$. The success of this approach led us to also try running triplets of consecutive layers in parallel, but we found it to perform worse.


\section{Efficient Parallelization of Blocks}
\label{sec:efficiency}

\begin{figure*}[b]
\centering
\usetikzlibrary{backgrounds}

\begin{tikzpicture}[scale=0.7,
        box/.style={draw, rectangle, minimum width=1.5cm, minimum height=0.75cm},
        reducebox/.style={draw, rectangle, minimum width=0.3cm, minimum height=0.3cm},
        arrow/.style={->, thick},
        line/.style={-, thick},
    ]
    \draw[->, thick] (2.4, 3.2) -- node[above left] {GPU} (3, 3.8);
    
    \node[rectangle, fill=green!30, minimum width=0.5cm, minimum height=0.3cm] at (7.5, 3) {};
    \node[right] at (8, 3) {Layer $k$};
    \node[rectangle, fill=yellow!30, minimum width=0.5cm, minimum height=0.3cm] at (7.5, 2.6) {};
    \node[right] at (8, 2.6) {Layer $k+1$};
    
    \node[box] (input) at (0, 0.55) {$x\in\mathbb{R}^{T\times D}$};

    \stackedtworectangles[3,-2]{V}{green!30}{yellow!30}{V};
    \stackedtworectangles[3,0]{K}{green!30}{yellow!30}{K};
    \stackedtworectangles[3,2]{Q}{green!30}{yellow!30}{Q};
    \draw[arrow] (input) -- +(1.5,0) |- (Q-E2left);
    \draw[arrow] (input) -- +(1.5,0) |- (K-E2left);
    \draw[arrow] (input) -- +(1.5,0) |- (V-E2left);

    \begin{scope}[on background layer]
        \draw[arrow] (input) -- +(1.7,0) |- (Q-E1left);
        \draw[arrow] (input) -- +(1.7,0) |- (K-E1left);
        \draw[arrow] (input) -- +(1.7,0) |- (V-E1left);
    \end{scope}

    \draw[line] (5.6, 0.55) -- (6.5, 0.55);    
    \draw[line] (5.6, 0.75) -- (6.5, 0.75);

    \stackedroundedtworectangles[6.25, -0.25]{mha}{white}{white}{Self att.};

    \stackedtworectangles[9, 0]{att}{green!30}{yellow!30}{att};
    \draw[line] (Q-E2right) -- +(0.75, 0) |- (mha-E2left);
    \draw[line] (K-E2right) --  (mha-E2left);
    \draw[line] (V-E2right) -- +(0.75, 0) |- (mha-E2left);

    \begin{scope}[on background layer]
        \draw[line] (Q-E1right) -- +(0.75, 0) |- (mha-E1left);
        \draw[line] (K-E1right) --  (mha-E1left);
        \draw[line] (V-E1right) -- +(0.75, 0) |- (mha-E1left);
    \end{scope}

    \draw[line] (mha-E2right) -- (att-E2left);
    \node[box, fill=green!30] (o1) at (13, 1.35) {$o_1\in\mathbb{R}^{T\times D}$};
    \node[box, fill=yellow!30] (o2) at (13, 0.15) {$o_2\in\mathbb{R}^{T\times D}$};
    \draw[arrow] (att-E2right) -- +(.6, 0) |- (o2);

    \begin{scope}[on background layer]
        \draw[arrow] (att-E1right) -- +(0.4, 0) |- (o1);
        \draw[arrow] (mha-E1right) -- (att-E1left);
    \end{scope}
    
    \node[reducebox] (allreduce) at (15.5, 0.75) {$+$};
    
    \draw[arrow] (o1) -- +(1.5,0.0) |- (allreduce);
    \draw[arrow] (o2) -- +(1.5,0.0) |- (allreduce);
    
    \node[box] (output) at (17.5, 0.75) {$o\in\mathbb{R}^{T\times D}$};
    \draw[arrow] (allreduce) -- (output);
\end{tikzpicture}
\caption{\label{fig:lp2} \textbf{LP attention implementation}. This diagram shows the implementation of the LP attention from Fig. \ref{fig:normal_and_pl_transformer_layer}b. The stacked layers represent different GPUs, the colors indicate different layers and the arrows express linear projections. In this case, the number of GPUs and the number of parallelized layers coincides and is two, which is the set-up that we use for all our experiments in this work. }
\end{figure*}

Naively trying to fuse two attention or MLP sub-blocks does not result
in a noticeable improvement of inference speed for large batch sizes and
sequence lengths, since in these situations, inference approaches the
the compute-bound regime. For this reason we focus our attention on the Tensor
Parallel setting, where each module's weights are split over two or more GPUs.
Rearranging the computational graph of two contiguous layers like in Fig.
\ref{fig:normal_and_pl_transformer_layer}b effectively reduces the number
of inter-GPU synchronizations by half. Now, each divergent path is computed
in parallel over multiple accelerators, and only the single intermediate and
final results need to be synchronized. While this approach is not numerically
equivalent to \eqref{eq:par}, we nonetheless---and quite surprisingly--- show
that it works well on already trained models, circumventing the need to train
from scratch.



\textbf{LP Multi-Head Attention.} Traditional tensor parallelism in MHA distributes attention
heads evenly across GPUs \citep{shoeybi2020megatronlmtrainingmultibillionparameter}, performing
self-attention and output projection locally before gathering results through an all-reduce
summation. Each GPU processes tensors of dimensions $Q,K,V,att\in\mathbb{R}^{T\times \frac{D}{g}}$,
where $T$ is sequence length, $D$ is feature dimension, and $g$ is the number of parallel workers.
The local output projection produces a low-rank $o_i\in\mathbb{R}^{T\times D}$ for each worker $i$,
and then a final all-reduce operation performs $o=\sum_i^g o_i$, computing the final output.

To implement LP, we increase the depth of the query, key, and value weight matrices
($W_Q,W_K,W_V\in\mathbb{R}^{(g_n\cdot h_d)\times D}$) and widen the output projection
($W_O\in\mathbb{R}^{D\times (n_h\cdot h_d)}$), where $n_h$ represents heads per GPU and $h_d$ is
head dimensionality. The reduction operation now will simultaneously compute the full-rank output
projections and the sum of all parallel layers (Fig.~\ref{fig:lp2}). Note that this approach requires
that $n_h$ is divisible by the total number of GPUs.


\textbf{LP FFN.} Standard tensor parallelism for single-hidden-layer FFNs splits the first layer's output across devices, generates low-rank outputs from the second layer, and sums them through reduction. To parallelize two FFN layers, we double the first layer's output dimensionality and perform separate output (low-rank) projections for each layer. A single reduction operation then serves the dual purpose of computing full outputs for each layer and combining their results, as shown in Fig.~ \ref{fig:normal_and_pl_transformer_layer}(b). In summary, LP for FFN just concatenates the up-projection weights and continues as normal TP, allowing for multiple GPUs to be allocated per parallelized layer.



\textbf{Handling of the LayerNorms.} Since we assume at least one GPU per parallelized layer,
we can assign each original LayerNorm to the divergent path that contains the attention
and FFN blocks from its original layer. We have also observed that using
the same merged LayerNorm with linear interpolation and spherical linear
interpolation in each divergent path yields good results. For the sake of
simplicity, we conduct all of our experiments using the original LayerNorms on
each divergent path.

\section{Experiments \& Results}
\label{sec:experiments}

\begin{figure*}[h!]
  \centering
  \begin{subfigure}[b]{0.4\linewidth}
  \includegraphics[width=\textwidth,clip]{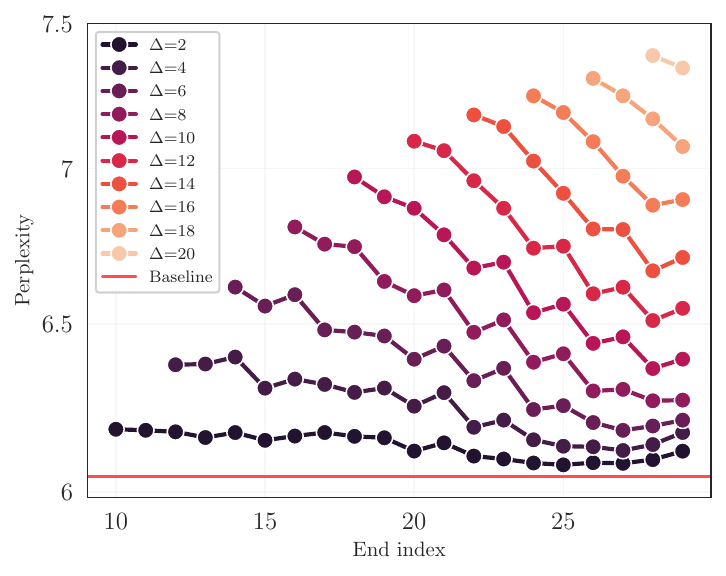}
  \caption{Llama2 7B.}
  \end{subfigure}
  \hspace{2em}
  \begin{subfigure}[b]{0.4\linewidth}
  \includegraphics[width=\textwidth,clip]{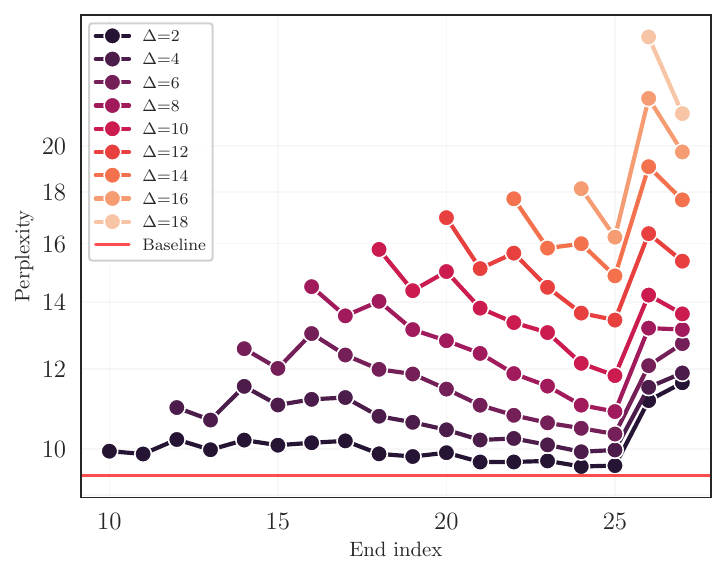}
  \caption{Llama3.2 3B.}
  \end{subfigure}
\vspace{-0.2cm}
\caption{\textbf{Perplexity when running pairs of consecutive layers in
parallel.} Perplexity of Llama2 7B and Llama3.2 3B models on the test set
of RedPajama\citep{together2023redpajama} when applying Layer Parallelism to
$\Delta$ consecutive layers. The parallelized interval for each data point is
$[\text{end index} - \Delta, \text{end index}]$, where \textit{end index} is the last
layer in the LLM to which LP was applied.
}
\label{fig:llama_ppl_sweep}
\end{figure*}

\begin{table*}[!hbt]
\centering
\setlength\tabcolsep{4pt}
\footnotesize
\caption{\label{tab:model-comparison}\textbf{5-shot In-Context Learning
accuracies across standard benchmarks}. Effective Depth shows the minimum number
of sequential operations from input to output after applying LP. We use the ablation on
Fig~\ref{fig:llama_ppl_sweep} to choose the LP configurations that minimized the perplexity.
*ifeval was evaluated on 0-shot performance.}
\begin{tabular}{c|c|cc|ccccccccc}
\toprule
Eff. Depth & Speed & Avg & Rel. & MMLU & PiQA & Arc E. & Arc C. & WinoG & OBQA & hswag & GSM8K & ifeval* \\

\midrule
\multicolumn{13}{c}{\textbf{Llama 2 7B (Chat)}}\\
\midrule

32  (base)  & x1.00 & 53.80 & 1.00 & \valstd{47.27}{0.4} & \valstd{77.69}{1.0} & \valstd{79.63}{0.8} & \valstd{49.06}{1.5} & \valstd{72.06}{1.3} & \valstd{33.00}{2.1} & \valstd{58.87}{0.5} & \valstd{22.97}{1.2} & 43.65  \\ 
27  (ours) &  x1.15  & 53.89 & 1.00 &   \valstd{47.46}{0.4} & \valstd{77.09}{1.0} & \valstd{78.03}{0.9} & \valstd{48.81}{1.5} & \valstd{71.82}{1.3} & \valstd{33.80}{2.1} & \valstd{58.39}{0.5} & \valstd{22.21}{1.1} & 47.36  \\
26  (ours) &  x1.19  & 53.25 & 0.99 &   \valstd{47.24}{0.4} & \valstd{76.66}{1.0} & \valstd{77.48}{0.9} & \valstd{47.01}{1.5} & \valstd{71.51}{1.3} & \valstd{34.00}{2.1} & \valstd{57.79}{0.5} & \valstd{19.03}{1.1} & 48.56  \\
25  (ours) &  x1.23  & 52.33 & 0.97 &   \valstd{47.67}{0.4} & \valstd{76.33}{1.0} & \valstd{77.19}{0.9} & \valstd{46.50}{1.5} & \valstd{70.09}{1.3} & \valstd{33.20}{2.1} & \valstd{57.24}{0.5} & \valstd{14.78}{1.0} & 47.96  \\
24  (ours) &  x1.28 & 49.62 & 0.92 &  \valstd{45.47}{0.4} & \valstd{76.55}{1.0} & \valstd{75.67}{0.9} & \valstd{43.69}{1.5} & \valstd{67.88}{1.3} & \valstd{30.40}{2.1} & \valstd{55.82}{0.5} & \valstd{9.63}{0.8} & 41.49  \\
23  (ours) &  x1.31 & 47.71 & 0.89 &  \valstd{42.71}{0.4} & \valstd{75.24}{1.0} & \valstd{74.28}{0.9} & \valstd{41.13}{1.4} & \valstd{66.54}{1.3} & \valstd{30.80}{2.1} & \valstd{54.06}{0.5} & \valstd{6.97}{0.7} & 37.65  \\
\midrule
\multicolumn{13}{c}{\textbf{Llama 3.2 3B (Instruct)}} \\
\midrule
28  (base)  & x1.00 & 60.88  & 1.00 & \valstd{59.56}{0.4} & \valstd{77.09}{1.0} & \valstd{79.29}{0.8} & \valstd{46.50}{1.5} & \valstd{70.24}{1.3} & \valstd{30.80}{2.1} & \valstd{52.53}{0.5} & \valstd{64.75}{1.3} & 67.15 \\ 
24  (ours) &  x1.12 & 57.16  & 0.94 &  \valstd{59.14}{0.4} & \valstd{76.22}{1.0} & \valstd{76.60}{0.9} & \valstd{44.88}{1.5} & \valstd{70.09}{1.3} & \valstd{30.0}{2.1} & \valstd{51.31}{0.5} & \valstd{45.87}{1.4} & 60.31 \\
23  (ours) &  x1.15 & 55.44  & 0.91 &  \valstd{59.05}{0.4} & \valstd{75.46}{1.0} & \valstd{76.18}{0.9} & \valstd{44.71}{1.5} & \valstd{68.98}{1.3} & \valstd{27.80}{2.0} & \valstd{51.16}{0.5} & \valstd{35.71}{1.3} & 59.95 \\
22  (ours) &  x1.19 & 51.59  & 0.85 &  \valstd{56.12}{0.4} & \valstd{75.30}{1.0} & \valstd{75.04}{0.9} & \valstd{45.73}{1.5} & \valstd{66.77}{1.3} & \valstd{28.80}{2.0} & \valstd{50.88}{0.5} & \valstd{10.01}{0.8} & 55.64 \\
21  (ours) &  x1.23 & 48.03  & 0.79 &  \valstd{50.72}{0.4} & \valstd{74.70}{1.0} & \valstd{72.56}{0.9} & \valstd{40.19}{1.4} & \valstd{64.09}{1.4} & \valstd{28.40}{2.0} & \valstd{49.47}{0.5} & \valstd{3.11}{0.5} & 49.04 \\
20  (ours) &  x1.28 & 45.76  & 0.75 &  \valstd{40.70}{0.4} & \valstd{74.10}{1.0} & \valstd{70.75}{0.9} & \valstd{39.42}{1.4} & \valstd{62.75}{1.4} & \valstd{29.00}{2.0} & \valstd{47.55}{0.5} & \valstd{3.11}{0.5} & 44.48 \\
\midrule
\multicolumn{13}{c}{\textbf{Qwen3 4B (Instruct)}} \\
\midrule
36  (base)  & x1.00 & 63.72 & 1.00 & \valstd{70.16}{0.4} & \valstd{76.44}{1.0} & \valstd{84.76}{0.8} & \valstd{58.79}{1.5} & \valstd{66.30}{1.3} & \valstd{37.20}{2.0} & \valstd{52.77}{0.5} & \valstd{84.99}{1.0} & 42.09 \\ 
31  (ours) &  x1.13  & 57.72 & 0.91 &  \valstd{68.87}{0.4} & \valstd{74.59}{1.0} & \valstd{81.82}{0.8} & \valstd{53.67}{1.5} & \valstd{65.98}{1.3} & \valstd{35.20}{2.2} & \valstd{50.10}{0.5} & \valstd{53.75}{1.0} & 35.49 \\
30  (ours) &  x1.15  & 55.45 & 0.87 &  \valstd{67.49}{0.4} & \valstd{74.97}{1.0} & \valstd{81.84}{0.9} & \valstd{52.39}{1.4} & \valstd{65.11}{1.3} & \valstd{33.40}{2.1} & \valstd{48.95}{0.5} & \valstd{36.77}{0.9} & 38.13 \\
29  (ours) &  x1.18  & 51.84 & 0.81 &  \valstd{63.56}{0.4} & \valstd{74.70}{1.0} & \valstd{79.88}{0.9} & \valstd{50.43}{1.4} & \valstd{63.22}{1.4} & \valstd{33.00}{2.2} & \valstd{47.25}{0.5} & \valstd{17.97}{0.8} & 36.57 \\
28  (ours) &  x1.21  & 49.09 & 0.77 &  \valstd{53.95}{0.4} & \valstd{74.32}{1.0} & \valstd{79.29}{0.9} & \valstd{48.98}{1.6} & \valstd{62.75}{1.3} & \valstd{30.0}{2.1} & \valstd{45.24}{0.5} & \valstd{12.66}{0.5} & 34.65 \\
27  (ours) &  x1.25  & 44.68 & 0.70 &  \valstd{44.09}{0.4} & \valstd{72.96}{1.0} & \valstd{76.68}{0.9} & \valstd{44.54}{1.5} & \valstd{59.67}{1.4} & \valstd{29.60}{2.0} & \valstd{43.22}{0.5} & \valstd{3.56}{0.5} & 27.82 \\
\midrule
\multicolumn{13}{c}{\textbf{Qwen3 14B (Instruct)}} \\
\midrule
40  (base)  & x1.00 & 68.75 & 1.00 & \valstd{78.83}{0.4} & \valstd{80.90}{1.0} & \valstd{87.75}{0.8} & \valstd{66.21}{1.4} & \valstd{74.51}{1.3} & \valstd{40.40}{2.1} & \valstd{61.33}{0.5} & \valstd{82.26}{1.2} & 46.52 \\ 
35  (ours) &  x1.12  & 65.04 & 0.95 &  \valstd{77.92}{0.4} & \valstd{79.82}{1.0} & \valstd{86.95}{0.9} & \valstd{64.68}{1.5} & \valstd{73.56}{1.3} & \valstd{38.80}{2.0} & \valstd{58.25}{0.5} & \valstd{61.64}{1.1} & 43.76 \\
34  (ours) &  x1.15  & 66.29 & 0.96 &  \valstd{77.42}{0.4} & \valstd{78.89}{1.0} & \valstd{85.61}{0.9} & \valstd{62.71}{1.5} & \valstd{73.88}{1.3} & \valstd{37.80}{2.0} & \valstd{57.92}{0.5} & \valstd{75.36}{1.0} & 47.00 \\
33  (ours) &  x1.18  & 64.64 & 0.94 &  \valstd{75.88}{0.4} & \valstd{79.05}{1.0} & \valstd{85.61}{0.9} & \valstd{60.58}{1.4} & \valstd{72.69}{1.4} & \valstd{37.40}{2.2} & \valstd{56.93}{0.5} & \valstd{69.83}{1.0} & 43.76 \\
32  (ours) &  x1.21  & 62.84 & 0.91 &  \valstd{73.70}{0.4} & \valstd{78.89}{1.0} & \valstd{84.81}{0.9} & \valstd{59.73}{1.4} & \valstd{71.03}{1.4} & \valstd{36.80}{2.0} & \valstd{55.75}{0.5} & \valstd{64.67}{0.9} & 40.17 \\
31  (ours) &  x1.24  & 60.11 & 0.87 &  \valstd{71.56}{0.4} & \valstd{78.56}{1.0} & \valstd{85.52}{0.9} & \valstd{59.98}{1.5} & \valstd{68.59}{1.3} & \valstd{36.40}{2.0} & \valstd{54.12}{0.5} & \valstd{48.22}{0.8} & 38.01 \\
30  (ours) &  x1.27  & 55.18 & 0.80 &  \valstd{65.43}{0.4} & \valstd{77.58}{1.0} & \valstd{83.84}{0.9} & \valstd{55.89}{1.5} & \valstd{67.32}{1.3} & \valstd{34.60}{2.1} & \valstd{51.97}{0.5} & \valstd{23.43}{0.8} & 36.57 \\
\bottomrule
\end{tabular}
\end{table*}

\begin{table}[h]
\centering
\footnotesize
\caption{\label{tab:finetuning} Benchmark accuracy restoration for LP-applied Qwen3 models. Fine-tuned entries use 4096 additional training steps; left columns report the fine-tuned accuracy, and right columns report the no-finetune/baseline values.}
\begin{tabular}{lc|ccc|ccc}
\toprule
\multirow{2}{*}{Model} & \multirow{2}{*}{Eff. depth} & \multicolumn{3}{c|}{Fine-tuned} & \multicolumn{3}{c}{No fine-tuning} \\
\cmidrule(lr){3-5}\cmidrule(lr){6-8}
 &  & MMLU & Arc C. & GSM-8K (\%) & MMLU & Arc C. & GSM-8K (\%) \\
\midrule
\multirow{4}{*}{Qwen3 4B} & 36 (Baseline) & --- & --- & ---             & \valstd{70.16}{0.4} & \valstd{58.79}{1.5} & \valstd{84.99}{1.0} \\
 & 31 & \valstd{69.11}{0.4} & \valstd{54.10}{1.5} & \valstd{62.47}{1.3} & \valstd{68.87}{0.4} & \valstd{53.67}{1.5} & \valstd{53.75}{1.0} \\
 & 30 & \valstd{68.43}{0.4} & \valstd{54.10}{1.5} & \valstd{56.56}{1.4} & \valstd{67.49}{0.4} & \valstd{50.43}{1.4} & \valstd{36.77}{0.9} \\
 & 27 & \valstd{61.96}{0.4} & \valstd{51.02}{1.6} & \valstd{48.29}{1.4} & \valstd{44.09}{0.4} & \valstd{44.54}{1.5} & \valstd{3.56}{0.5} \\
\midrule
\multirow{3}{*}{Qwen3 14B} & 40 (Baseline) & --- & --- & ---            & \valstd{78.83}{0.4} & \valstd{66.21}{1.4} & \valstd{82.26}{1.2} \\
 & 35 & \valstd{77.89}{0.3} & \valstd{63.05}{1.4} & \valstd{81.73}{1.1} & \valstd{77.92}{0.4} & \valstd{64.68}{1.5} & \valstd{61.64}{1.1} \\
 & 32 & \valstd{74.38}{0.4} & \valstd{59.30}{1.4} & \valstd{71.27}{1.3} & \valstd{73.70}{0.3} & \valstd{59.73}{1.4} & \valstd{64.67}{0.9} \\
\bottomrule
\end{tabular}
\end{table}

\begin{figure*}[h]
\centering
\includegraphics[width=0.9\textwidth]{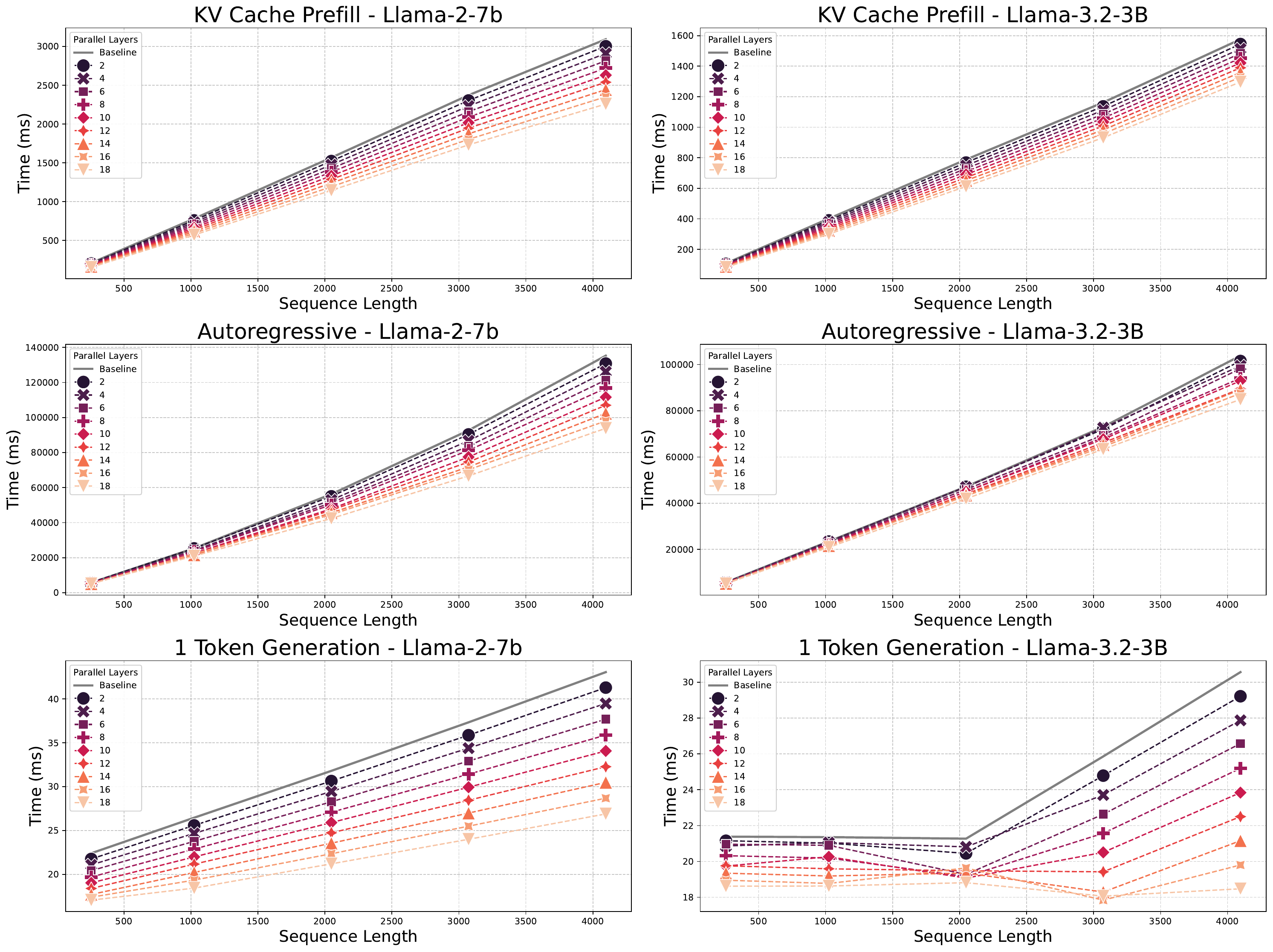}
\caption{\label{fig:tp_avg_time} \textbf{Wall clock time to complete the following inference tasks}: KV Cache pre-filling, autoregressive generation, and single token generation with a pre-filled KV Cache. $\Delta$ indicates how many layers have been merged using LP (e.g. a $\Delta$ of 4 indicates that 2 groups of 2 layers have been converted to 2 effective layers). The gains in inference speed are roughly proportional to the amount of LP. The 1-token generation task for Llama 3.2 3B does not saturate the GPU compute until a sequence length of 2048. Even in this regime, LP benefits from considerable speed-ups.}
\end{figure*}

In this section, we evaluate Layer Parallelism across three dimensions: inference speed improvements, impact on In-Context Learning performance, and the potential to recover model accuracy through targeted fine-tuning of parallelized layers.

\textbf{Experimental protocol.} For all our experiments, we use a node with
x2 A100 SXM4 80Gb GPUs, x4 AMD EPYC 7742 CPUs, and 512Gb of RAM. We test for
varying sequence lengths, up to $4096$ (Llama's context window), with a batch
size of $1$ unless indicated otherwise. We consider two models of the Llama
family: Llama2 7B, and Llama3.2 3B, as well as two sizes from Qwen3: 4B and
14B. Given a desired effective depth, we replace the required number of normal
layers with LP layers. For Llama 2 7B and 3.2 3B, the LP layers are selected
based on the configuration that minimized the PPL for a given amount of LP
(Fig. \ref{fig:llama_ppl_sweep}). For Qwen3, LP is applied until the 4th to last
decoder layer. The rest of the layers implement the tensor parallel approach
as described in \citep{shoeybi2020megatronlmtrainingmultibillionparameter}.
For evaluation, we measure the ICL 5-shot accuracies using the \verb|lm-eval|
package \citep{eval-harness}. We test the ICL accuracy of the models on
several tasks: MMLU \citep{hendrycks2021measuringmassivemultitasklanguage},
PiQA \citep{bisk2019piqareasoningphysicalcommonsense}, ARC Easy, ARC
Challenge, Winogrande \citep{sakaguchi2021winogrande}, OpenBookQA
\citep{mihaylov2018suitarmorconductelectricity}, Hellaswag
\citep{zellers2019hellaswag}, GSM-8K \citep{cobbe2021trainingverifierssolvemath}
and ifeval \citep{zhou2023instructionfollowingevaluationlargelanguage}. The
perplexity of the models is always evaluated against a subset of the test set of
RedPajama \citep{together2023redpajama}.


\textbf{Impact of LP on PPL and ICL accuracies.} We first examine how perplexity evolves when applying LP across layer sequences of different lengths and depths. Fig. \ref{fig:llama_ppl_sweep} reveals a common optimal sequence end-index minimizing perplexity, found at layers 28 for Llama2 7B and 25 for Llama3.2 3B. Table \ref{tab:model-comparison} compares the In-Context Learning performance across models with varying effective depths. Performance declines gradually as LP increases, followed by a sharp drop beyond a certain threshold. Specifically, this occurs after reducing effective depth by 9 layers for Qwen3 14B, by 7 layers for Llama2 7B and Qwen3 4B, and by 5 layers for Llama3.2 3B. These results indicate that larger models are more robust to the computational graph modifications from LP, suggesting that our approach is likely applicable to current commercial-scale LLMs used in major deployments.

It is worth noting that, unlike the other benchmarks, GSM-8K
already drops severely in performance when applying low amounts of
LP. Recent mechanistic interpretability research shows that LLMs
have special circuitry for math operations, localized in a small
set of parameters \citep{stolfo2023mechanistic,yu2024interpreting}.
\citep{christ2025mathneurosurgeryisolatinglanguage} identify math specific
parameters, and report a drop in accuracy of 17\% when pruning them. We hypothesize
that the changes in the computational graph by the use of LP in some of the late
layers of the LLM interfere with these fragile and sparse subnetworks, while
leaving general language competence largely intact.


\textbf{Impact on the inference speed.} We run an ablation over several
configurations and input sequence lengths on Figure \ref{fig:tp_avg_time} to
test the speed on three different tasks: KV-Cache pre-filling, autoregressive
generation up to the sequence length(with KV-Cache) and 1-token generation with
a pre-filled KV-Cache of the corresponding sequence length. Our ablations show
that the speed gain is strongly correlated with the reduction of the effective
depth of the model. For the effective depths of 25 ($\Delta=14$) in Llama 2 7B,
we observe an average speed-up of 1.29x at the largest sequence length in the
1-token generation task. Likewise, for an effective depth of 23 ($\Delta=10$)
in Llama 3.2 3B, we report a speed-up of 1.22x. For more aggressive parallelism,
$\Delta=18$ and $\Delta=16$, we report a speed-up of 1.38x and 1.35x , at the
expense of a large drop in ICL accuracy.

\textbf{Fine-tuning for performance recovery.} While LP provides speed improvements, associated
architectural modifications may degrade model performance. To counteract this, we explored whether
fine-tuning could effectively restore the original model's capabilities. We apply LP to some
configurations of Qwen3-4B and Qwen3-14B (Table~\ref{tab:finetuning}), and fine-tune the LP
layers on randomly selected samples from the RedPajama training set \citep{together2023redpajama}.
We employ a batch size of 32, a linear learning rate schedule starting at $1e{-4}$ and the AdamW
optimizer \citep{loshchilov2017decoupled}. We observe a significant restoration of the benchmark
accuracies for Qwen3-4B with an effective depth of 27, especially on GSM-8K, which recovered from
near-zero levels. Less aggressive usage of LP results in a less pronounced recovery of the accuracy,
and fails to fully recover the original model's performance. It is possible that additional fine-tuning,
or smarter tuning strategies, could yield further improvements, but resource constraints limited the scope
of our experiments.

\section{Limitations}
\label{sec:limitations}





The effectiveness of our approach exhibits notable variations across
model scales. Smaller models show reduced benefits, likely due to their less
sparse activation patterns and more tightly coupled layer dependencies. This
degradation becomes more pronounced as the LP sequence length increases, suggesting
a practical upper limit to the number of layer pairs that can be effectively
parallelized.

Regarding the fine-tuning, while some performance loss can be mitigated, we were
unable to fully recover the baseline model's performance levels. This suggests
fundamental trade-offs between computational efficiency and model capability
that cannot be entirely eliminated through optimization, or that more involved
fine-tuning strategies might be required.

Moreover, determining the 'true' effective depth—the optimal configuration of
parallel layer pairs—remains an open challenge as there is no theoretical
framework for predicting the optimal grouping strategy.

These limitations highlight important directions for future research,
particularly in developing more robust methods for determining optimal layer
groupings and investigating the interplay between our approach and other
efficiency-oriented techniques.


\section{Conclusion}




In this work, we presented Layer Parallelism, a novel approach that exploits
independence patterns between transformer layers to optimize LLM inference.
By restructuring the computational graph to enable parallel execution of
consecutive layer pairs through tensor parallelism, we achieved substantial
speed improvements without model retraining. Our method reduced the effective
depth of Llama 2 7B by 21\% while maintaining 98\% of the original performance
(without fine-tuning), yielding up to a 1.29x improvement in inference speed
for single-token generation with long sequences. Moreover, we show that we can
recover some of the lost accuracy through naive fine-tuning.

These results challenge the conventional view that transformer layers must
process information strictly sequentially, suggesting instead that certain
layers can operate independently without significant performance loss. From a
practical standpoint, LP offers a straightforward approach to improve inference
efficiency in production environments. Future work could focus on developing
theoretical frameworks to predict optimal layer groupings, investigating
interactions with other efficiency techniques such as quantization, and
understanding the fundamental principles behind layer independence. Despite its
limitations, LP represents a practical advancement in making LLM deployment more
efficient and economically viable.




\section*{Acknowledgments}

We would like to thank Benjamin Rio for the thoughtful discussions and help running some of the performance experiments. Ram\'on's research is supported by META.
 
\bibliographystyle{tmlr}
\bibliography{bibliography}

\appendix

\newpage
\appendix
\section{Theoretical Analysis of Layer Parallelism}
\label{app:lp_theory}

This section provides a theoretical justification for Layer Parallelism (LP), analyzing the approximation error introduced by the LP computational graph and connecting it to the empirical observations in the main paper.

\subsection{Sequential vs.\ Layer-Parallel Computation}

Consider two consecutive transformer decoder layers $\ell_k$ and $\ell_{k+1}$ in a pre-norm architecture. Let $A_k(\cdot)$ denote the attention residual and $F_k(\cdot)$ the feed-forward residual of layer $k$. 

\paragraph{Exact sequential computation.} The standard forward pass computes:
\begin{align}
u_k &= x + A_k(x), \label{eq:seq1}\\
h_k &= u_k + F_k(u_k), \label{eq:seq2}\\
u_{k+1} &= h_k + A_{k+1}(h_k), \label{eq:seq3}\\
h_{k+1} &= u_{k+1} + F_{k+1}(u_{k+1}), \label{eq:seq4}
\end{align}
where $h_{k+1} = T_{\mathrm{seq}}(x)$ is the two-layer sequential output.

\paragraph{Layer Parallelism computation.} LP evaluates both attention modules at the shared input $x$, combines their outputs, and feeds the result to both FFN modules:
\begin{align}
\tilde{u} &= x + A_k(x) + A_{k+1}(x), \label{eq:lp1}\\
T_{\mathrm{LP}}(x) &= \tilde{u} + F_k(\tilde{u}) + F_{k+1}(\tilde{u}). \label{eq:lp2}
\end{align}
This matches the (LP) equation in \S\ref{sec:effective-depth} and the implementation in Fig.~\ref{fig:normal_and_pl_transformer_layer}(b), where divergent paths share intermediate states.

\subsection{First-Order Error Analysis}

The approximation error $\mathcal{E}(x) = T_{\mathrm{seq}}(x) - T_{\mathrm{LP}}(x)$ arises from evaluating submodules at different inputs. We decompose this error into three components.

\paragraph{Component 1: Attention evaluation shift.}
In the sequential computation, $A_{k+1}$ is evaluated at $h_k = x + A_k(x) + F_k(u_k)$, whereas LP evaluates it at $x$. Defining $\Delta_1 = h_k - x = A_k(x) + F_k(u_k)$, a first-order Taylor expansion yields:
\begin{equation}
A_{k+1}(h_k) - A_{k+1}(x) \approx J_{A_{k+1}}(x)\, \Delta_1,
\label{eq:err_attn}
\end{equation}
where $J_{A_{k+1}}(x)$ is the Jacobian of $A_{k+1}$ at $x$.

\paragraph{Component 2: First FFN evaluation shift.}
In the sequential computation, $F_k$ is evaluated at $u_k = x + A_k(x)$, whereas LP evaluates it at $\tilde{u} = x + A_k(x) + A_{k+1}(x)$. Since $\tilde{u} - u_k = A_{k+1}(x)$:
\begin{equation}
F_k(u_k) - F_k(\tilde{u}) \approx -J_{F_k}(u_k)\, A_{k+1}(x).
\label{eq:err_ffn1}
\end{equation}

\paragraph{Component 3: Second FFN evaluation shift.}
In the sequential computation, $F_{k+1}$ is evaluated at $u_{k+1} = h_k + A_{k+1}(h_k)$, whereas LP evaluates it at $\tilde{u}$. The difference is:
\begin{equation}
u_{k+1} - \tilde{u} = F_k(u_k) + \bigl[A_{k+1}(h_k) - A_{k+1}(x)\bigr].
\label{eq:uk1_diff}
\end{equation}
Substituting the first-order approximation from \eqref{eq:err_attn}:
\begin{equation}
F_{k+1}(u_{k+1}) - F_{k+1}(\tilde{u}) \approx J_{F_{k+1}}(\tilde{u})\bigl[F_k(u_k) + J_{A_{k+1}}(x)\,\Delta_1\bigr].
\label{eq:err_ffn2}
\end{equation}

\paragraph{Total error bound.}
Combining equations \eqref{eq:err_attn}--\eqref{eq:err_ffn2} and taking norms:
\begin{equation}
\boxed{
\|\mathcal{E}(x)\| \lesssim 
\underbrace{\|J_{A_{k+1}}\|\,\|\Delta_1\|}_{\text{attention shift}}
+ \underbrace{\|J_{F_k}\|\,\|A_{k+1}(x)\|}_{\text{FFN}_k\text{ shift}}
+ \underbrace{\|J_{F_{k+1}}\|\bigl(\|F_k(u_k)\| + \|J_{A_{k+1}}\|\,\|\Delta_1\|\bigr)}_{\text{FFN}_{k+1}\text{ shift}},
}
\label{eq:error_bound}
\end{equation}
where Jacobian norms are operator norms evaluated at the appropriate inputs (suppressed for clarity).

\subsection{Implications for Layer Selection}
\label{sec:theory_implications}

The error bound \eqref{eq:error_bound} reveals when LP introduces minimal degradation:

\paragraph{Small residual updates favor LP.} 
The bound depends on $\|\Delta_1\| = \|A_k(x) + F_k(u_k)\|$, $\|A_{k+1}(x)\|$, and $\|F_k(u_k)\|$. Layers where attention and FFN residuals are small relative to the residual stream contribute less error. This is consistent with the ``residual stream dominance'' phenomenon observed in deep transformers.

\paragraph{Low Jacobian sensitivity favors LP.}
The terms $\|J_{A_{k+1}}\|$ and $\|J_{F_k}\|$, $\|J_{F_{k+1}}\|$ measure how sensitive each submodule is to input perturbations. Layers that are relatively insensitive to their exact input introduce less LP error.

\paragraph{Connection to CKA analysis (Fig.~\ref{fig:cka}).}
The CKA similarity between standard and counterfactual activations (with residual removed) serves as an empirical proxy for the Jacobian sensitivity. High CKA similarity indicates that the module output is relatively invariant to the upstream residual---precisely the condition under which the Jacobian terms in \eqref{eq:error_bound} are effectively small in the directions that matter. The plateau of high CKA values in mid-to-late layers (Fig.~\ref{fig:cka}) corresponds to the region where LP is most effective (Fig.~\ref{fig:matrix-2parallel}).

\paragraph{Error accumulation through the network.}
Errors injected at layer $k$ propagate through all subsequent layers via the Jacobian chain:
\begin{equation}
\|\mathcal{E}_{\mathrm{output}}\| \lesssim \|\mathcal{E}_k\| \prod_{\ell > k} \|I + J_{f_\ell}\|,
\label{eq:accumulation}
\end{equation}
where $f_\ell$ is the full residual map of layer $\ell$. This implies:
\begin{itemize}[nosep]
    \item \textbf{Avoid early layers}: Errors introduced early are amplified by many subsequent Jacobians.
    \item \textbf{Preserve a sequential tail}: The final layers before the output logits are typically most sensitive (the language modeling head amplifies perturbations), so leaving them sequential stabilizes the output distribution.
\end{itemize}

\subsection{Why Contiguous 2-Parallel Works Best}

The ``2-parallel'' scheme (parallelizing consecutive pairs rather than arbitrary groups) succeeds because:

\begin{enumerate}[nosep]
    \item \textbf{Local ordering matters less than global ordering.} Shuffling experiments (Fig.~\ref{fig:matrix-shuffling}) show that adjacent layers are more interchangeable than distant ones, likely because they operate at similar levels of abstraction.
    
    \item \textbf{Error terms remain bounded.} Within a single LP pair, the error is first-order in residual magnitudes. Chaining $n$ LP pairs gives $n$ independent first-order errors rather than a single large error from parallelizing all $2n$ layers simultaneously.
    
    \item \textbf{Intermediate synchronization corrects drift.} Between LP pairs, the outputs are summed and re-normalized, preventing error accumulation within the LP region.
\end{enumerate}

This explains why parallelizing triplets performs worse (as noted in \S\ref{sec:effective-depth}): the second-order cross-terms become significant, and there is no intermediate synchronization to correct the trajectory.

\subsection{Connection to GSM-8K Degradation}

The disproportionate drop in GSM-8K accuracy under LP (Table~\ref{tab:model-comparison}) is consistent with recent findings that mathematical reasoning relies on sparse, localized circuits \citep{stolfo2023mechanistic, christ2025mathneurosurgeryisolatinglanguage}. These circuits likely have:
\begin{itemize}[nosep]
    \item Larger effective Jacobians in the relevant directions (high sensitivity to precise intermediate states).
    \item Less redundancy, so the LP approximation error is not absorbed by parallel pathways.
\end{itemize}
General language competence, by contrast, is distributed across many redundant pathways and is therefore more robust to the input perturbations introduced by LP.



\newpage
\section{Comparison with Other Tensor Parallelism Optimizations}
\label{app:comparison}

In this section, we compare Layer Parallelism (LP) with recent methods that aim to reduce communication overhead in tensor-parallel LLM inference: Sync-Point Drop (SPD)~\citep{kim2025spdsyncpointdropefficient}, selective low-bit communication~\citep{dong2024lowbitcommunicationtensorparallel}, and microscaling (MX) format compression~\citep{hansenpalmus2024communicationcompressiontensorparallel}.

\subsection{Method Overview}

\paragraph{Sync-Point Drop (SPD).} \citet{kim2025spdsyncpointdropefficient} selectively removes the all-reduce synchronization after the self-attention output projection, retaining only the FFN synchronization per block. They introduce modified block designs to minimize information loss and classify blocks into three sensitivity categories (in-sensitive, sensitive, extremely sensitive), applying block-to-block distillation to recover accuracy in sensitive layers.

\paragraph{Selective Low-bit Communication.} \citet{dong2024lowbitcommunicationtensorparallel} compress communicated activations by quantizing most features to INT4 while keeping a small fraction ($1/64$) of high-range outlier features in BF16. This reduces communication from 16 bits to $\sim$4.2 bits per value on average, preserving outlier information critical for model performance.

\paragraph{MX Format Compression.} \citet{hansenpalmus2024communicationcompressiontensorparallel} apply microscaling (MX) quantization formats (FP4/FP5 with block-wise scaling) to compress activations before inter-device communication, achieving 3.5--4.5$\times$ compression ratios.

\paragraph{Layer Parallelism (Ours).} LP restructures the computational graph to execute consecutive layer pairs in parallel, reducing the number of sequential synchronization points. Unlike quantization-based methods, LP modifies the \emph{computation order} rather than the \emph{communication encoding}.

\subsection{Quantitative Comparison}

Table~\ref{tab:method-comparison} compares results across methods. Direct comparison is challenging due to different models, hardware configurations, and evaluation metrics; we match the closest available configurations.

\begin{table}[h]
\centering
\caption{Comparison of tensor parallelism optimization methods. Speedup is relative to standard tensor parallelism. For accuracy: SPD and LP report average zero-shot accuracy change; Low-bit reports performance retention; MX reports perplexity increase. Results marked $^\dagger$ are at 70\% SPD application; $^\ddagger$ indicates hardware-dependent results on PCIe-connected GPUs (L4); $^\S$ indicates NVLink-connected GPUs (A100).}
\label{tab:method-comparison}
\small
\setlength{\tabcolsep}{3pt}
\begin{tabular}{llcccc}
\toprule
\textbf{Method} & \textbf{Model} & \textbf{GPUs} & \textbf{Speedup} & \textbf{Accuracy Impact} & \textbf{Mechanism} \\
\midrule
\multicolumn{6}{c}{\textit{7B-scale Models}} \\
\midrule
SPD$^\dagger$ & LLaMA2-7B & 8 & 1.10$\times$ & $-$1.0\% avg acc & Drop attn sync \\
Low-bit & LLaMA2-13B & 8 & --- & 99.5\% retained & Quant.\ comm \\
MX Compress & LLaMA2-7B & 2$^\ddagger$ & 1.03$\times$ & +3.2\% PPL & Quant.\ comm \\
\textbf{LP (Ours)} & LLaMA3.2-3B & 2 & \textbf{1.19$\times$} & $-$2.5\% avg acc & Parallel layers \\
\midrule
\multicolumn{6}{c}{\textit{13B-scale Models}} \\
\midrule
SPD$^\dagger$ & LLaMA2-13B & 8 & 1.12$\times$ & $-$1.0\% avg acc & Drop attn sync \\
Low-bit & LLaMA2-13B & 8 & --- & 99.5\% retained & Quant.\ comm \\
MX Compress & LLaMA2-13B & 4$^\ddagger$ & 2.05$\times$ & +3.2\% PPL & Quant.\ comm \\
\textbf{LP (Ours)} & Qwen3-14B & 2 & \textbf{1.15$\times$} & $-$4.0\% avg acc & Parallel layers \\
\midrule
\multicolumn{6}{c}{\textit{70B-scale Models}} \\
\midrule
SPD$^\dagger$ & LLaMA2-70B & 8 & 1.20$\times$ & $-$0.9\% avg acc & Drop attn sync \\
MX Compress & LLaMA2-70B & 8$^\ddagger$ & 1.83--2.08$\times$ & +1.7\% PPL & Quant.\ comm \\
MX Compress & LLaMA2-70B & 4$^\S$ & 0.56--0.70$\times$ & +1.7\% PPL & Quant.\ comm \\
\bottomrule
\end{tabular}
\end{table}

Table~\ref{tab:accuracy-comparison} provides a detailed accuracy comparison on common benchmarks where available.

\begin{table}[h]
\centering
\caption{Accuracy comparison on zero-shot benchmarks. Values show absolute accuracy (\%) or relative change from baseline. SPD results use ZS+B2B configuration at 70\% SPD; Low-bit uses INT4+Selected BF16 at 4.2 bits.}
\label{tab:accuracy-comparison}
\small
\setlength{\tabcolsep}{4pt}
\begin{tabular}{llccccc}
\toprule
\textbf{Method} & \textbf{Model} & \textbf{ARC-e} & \textbf{ARC-c} & \textbf{HellaSwag} & \textbf{WinoGrande} & \textbf{Avg} \\
\midrule
\multicolumn{7}{c}{\textit{Baseline (no optimization)}} \\
\midrule
--- & LLaMA2-13B & 79.5 & 48.7 & 60.0 & 72.2 & 65.1 \\
--- & Gemma 2 27B & 87.7 & 62.4 & 65.4 & 79.1 & 73.7 \\
\midrule
\multicolumn{7}{c}{\textit{Low-bit Communication~\citep{dong2024lowbitcommunicationtensorparallel}}} \\
\midrule
INT4+Sel.\ BF16 & LLaMA2-13B & 79.1 & 47.4 & 59.5 & 72.9 & 64.7 \\
& & \textcolor{gray}{($-$0.5\%)} & \textcolor{gray}{($-$2.8\%)} & \textcolor{gray}{($-$0.9\%)} & \textcolor{gray}{(+1.0\%)} & \textcolor{gray}{($-$0.6\%)} \\
INT4+Sel.\ BF16 & Gemma 2 27B & 86.5 & 61.0 & 63.9 & 76.5 & 72.0 \\
& & \textcolor{gray}{($-$1.4\%)} & \textcolor{gray}{($-$2.2\%)} & \textcolor{gray}{($-$2.3\%)} & \textcolor{gray}{($-$3.3\%)} & \textcolor{gray}{($-$2.3\%)} \\
\midrule
\multicolumn{7}{c}{\textit{SPD~\citep{kim2025spdsyncpointdropefficient} at 70\% blocks}} \\
\midrule
ZS+B2B & LLaMA2-13B (8-GPU) & \multicolumn{4}{c}{---} & $\sim$65\% \\
& & \multicolumn{4}{c}{} & \textcolor{gray}{($<$1\% drop)} \\
ZS+B2B & LLaMA2-70B (8-GPU) & \multicolumn{4}{c}{---} & $\sim$66\% \\
& & \multicolumn{4}{c}{} & \textcolor{gray}{($-$0.9\%)} \\
\bottomrule
\end{tabular}
\end{table}

\subsection{Key Observations}

\paragraph{Complementary approaches.} The methods target different bottlenecks: LP and SPD reduce the \emph{number} of synchronization points, while low-bit and MX compression reduce the \emph{size} of each synchronization. These are largely orthogonal and could be combined---for instance, applying LP to reduce sync-points by 50\%, then using INT4 quantization to compress remaining communications by 4$\times$.

\paragraph{Hardware sensitivity.} Communication compression methods show strong hardware dependence. \citet{hansenpalmus2024communicationcompressiontensorparallel} report 2$\times$ TTFT speedup on PCIe-connected L4 GPUs (64 GB/s bandwidth) but \emph{slowdown} on NVLink-connected A100s (600 GB/s bandwidth) due to quantization overhead exceeding communication savings. In contrast, LP and SPD benefit from reduced sync-points regardless of interconnect speed, though gains are more pronounced on slower interconnects.

\paragraph{Model size scaling.} Both SPD and LP show improved robustness on larger models:
\begin{itemize}[nosep]
    \item \textbf{SPD}: 44\% of blocks are in-sensitive in LLaMA2-7B vs.\ 75\% in LLaMA2-70B~\citep{kim2025spdsyncpointdropefficient}
    \item \textbf{LP}: Larger models (Qwen3-14B) tolerate greater effective depth reduction than smaller models (LLaMA3.2-3B) before sharp accuracy drops (see Table~\ref{tab:model-comparison})
\end{itemize}

\paragraph{Trade-off characteristics.} Each method exhibits distinct accuracy--speed trade-offs:
\begin{itemize}[nosep]
    \item \textbf{SPD}: Graceful degradation; requires per-block sensitivity analysis and optional distillation
    \item \textbf{Low-bit}: Minimal degradation ($\sim$0.5--2\%) by preserving outlier features; no speedup measured
    \item \textbf{MX Compress}: Perplexity increases 1--3\%; speedup highly hardware-dependent
    \item \textbf{LP (Ours)}: Uniform accuracy drop across layers; GSM-8K disproportionately affected; fine-tuning recovers $\sim$50\% of loss
\end{itemize}

\subsection{Combining LP with Communication Compression}

A promising direction is combining LP with quantization-based compression. Table~\ref{tab:combined-estimate} estimates potential combined benefits.

\begin{table}[h]
\centering
\caption{Estimated combined speedup from LP + communication quantization on bandwidth-constrained hardware. LP reduces sync-points; INT4/MX4 reduces per-sync data volume by $\sim$4$\times$.}
\label{tab:combined-estimate}
\small
\begin{tabular}{lccc}
\toprule
\textbf{Optimization} & \textbf{Sync Reduction} & \textbf{Comm.\ Compression} & \textbf{Est.\ Speedup} \\
\midrule
Baseline (TP) & 0\% & 1$\times$ & 1.0$\times$ \\
LP only & 25--50\% & 1$\times$ & 1.15--1.25$\times$ \\
INT4/MX4 only & 0\% & 4$\times$ & 1.5--2.0$\times$ \\
\textbf{LP + INT4/MX4} & 25--50\% & 4$\times$ & \textbf{1.8--2.5$\times$} \\
\bottomrule
\end{tabular}
\end{table}

We leave empirical validation of combined LP + quantization to future work.

\newpage
\section{Ablation: Tokens per second}

\begin{figure*}[!h]
\centering
\includegraphics[width=\textwidth]{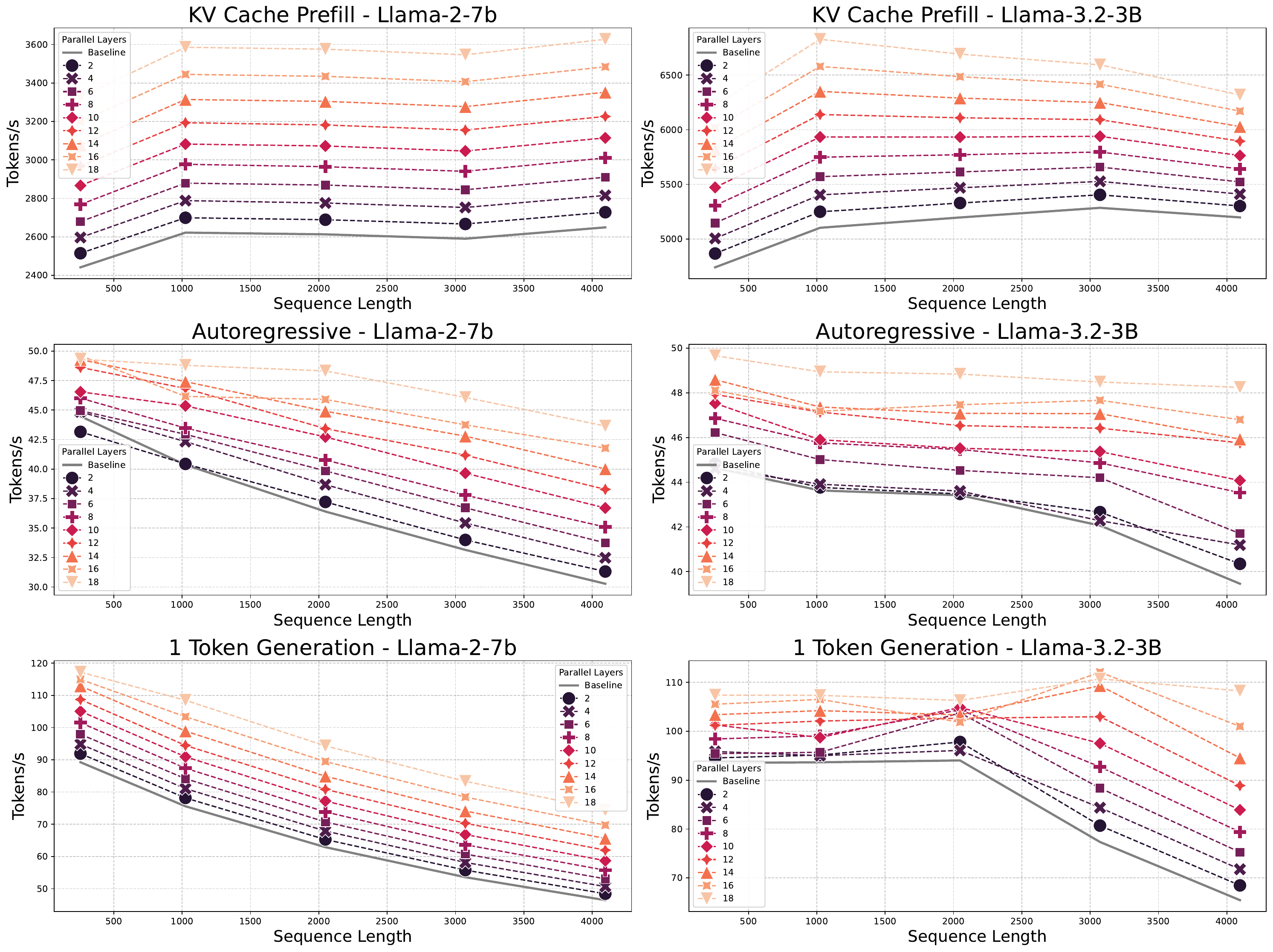}
\caption{\label{fig:tp_tot_tokens} Tokens per second when completing the following inference tasks: KV Cache pre-filling for a given sequence length, autoregressive generation up to the indicated sequence length, and single token generation with a pre-filled KV Cache of the indicated sequence length. The baseline is the original model with all layers making use of Tensor Parallelism. The Parallel Layers number ($\Delta$) indicates how many layers have been merged using Layer Parallelism (e.g. a $\Delta$ of 4 indicates that 2 groups of 2 layers have been converted to 2 effective layers). The number of tokens is computed as the sum of the input tokens and the output tokens for each forward pass.}
\end{figure*}



\newpage
\section{Generalization to multiple GPUs}

\begin{figure*}[!h]
\centering
\begin{tikzpicture}[scale=0.70,
        box/.style={draw, rectangle, minimum width=1.5cm, minimum height=0.75cm},
        reducebox/.style={draw, rectangle, minimum width=0.3cm, minimum height=0.3cm},
        arrow/.style={->, thick},
        line/.style={-, thick},
    ]
    \draw[->, thick] (2.2, 3) -- node[above, left] {GPU} (3, 3.8);
    
    \node[rectangle, fill=green!30, minimum width=0.5cm, minimum height=0.3cm] at (7.5, 3) {};
    \node[right] at (8, 3) {Layer $k$};
    \node[rectangle, fill=yellow!30, minimum width=0.5cm, minimum height=0.3cm] at (7.5, 2.6) {};
    \node[right] at (8, 2.6) {Layer $k+1$};
    
    \node[box] (input) at (0, 0.15) {$x\in\mathbb{R}^{T\times D}$};
    \stackedrectangles[3,-2]{V}{green!30}{yellow!30}{V};
    \stackedrectangles[3,0]{K}{green!30}{yellow!30}{K};
    \stackedrectangles[3,2]{Q}{green!30}{yellow!30}{Q};
    \draw[arrow] (input) -- +(1.5,0) |- (Q-E4left);
    \draw[arrow] (input) -- +(1.5,0) |- (K-E4left);
    \draw[arrow] (input) -- +(1.5,0) |- (V-E4left);
    \stackedroundedrectangles[6, 0]{mha}{white}{white}{Self att.};
    \draw[line] (Q-E4right) -- +(0.75, 0) |- (mha-E4left);
    \draw[line] (K-E4right) -- +(0.75, 0) |- (mha-E4left);
    \draw[line] (V-E4right) -- +(0.75, 0) |- (mha-E4left);
    \stackedrectangles[9, 0]{att}{green!30}{yellow!30}{att};
    \draw[arrow] (mha-E4right) -- (att-E4left);
    \node[box, fill=green!30] (o1) at (13, 2.25) {$o_1\in\mathbb{R}^{T\times D}$};
    \node[box, fill=green!30] (o2) at (13, 0.75) {$o_2\in\mathbb{R}^{T\times D}$};
    \node[box, fill=yellow!30] (o3) at (13, -0.75) {$o_3\in\mathbb{R}^{T\times D}$};
    \node[box, fill=yellow!30] (o4) at (13, -2.25) {$o_4\in\mathbb{R}^{T\times D}$};
    \draw[arrow] (att-E4right) -- +(.8, 0) |- (o4);
    \draw[arrow] (att-E3right) -- +(.8, 0) |- (o3);
    \draw[arrow] (att-E2right) -- +(0.6, 0) |- (o2);
    \draw[arrow] (att-E1right) -- +(0.2, 0) |- (o1);
    
    \node[reducebox] (allreduce) at (15.5, 0) {$+$};
    
    \draw[arrow] (o1) -- +(1.5,0) |- (allreduce);
    \draw[arrow] (o2) -- +(1.5,0) |- (allreduce);
    \draw[arrow] (o3) -- +(1.5,0) |- (allreduce);
    \draw[arrow] (o4) -- +(1.5,0) |- (allreduce);
    
    \node[box] (output) at (17.5, 0) {$o\in\mathbb{R}^{T\times D}$};
    \draw[arrow] (allreduce) -- (output);
\end{tikzpicture}
\caption{\label{fig:tplpcombo} Layer Parallelism in the case of parallelizing
two layers over four accelerators. The stacked layers represent the tensor
parallelism, and the colors indicate the processing of different previously
contiguous layers. $Q,K,V,\text{att} \in\mathbb{R}^{T\times \frac{2D}{g}}$,
where $D$ is the feature dimension and $g$ is the total number of accelerators.}
\end{figure*}

Layer Parallelism allows one to allocate $N\geq 1$ accelerators for each layer. The implementation remains the same, but now each layer is parallelized using tensor parallelism over its assigned accelerators. Note that both reduction operations (tensor parallel and layer parallel) are nicely executed with a single all-reduce call. 

To confirm that the proposed scheme scales to commodity 4-GPU servers, we benchmark Llama 2 7B on a node with $4\times$ NVIDIA A100 80\,GB PCIe accelerators while running the 4-GPU LP implementation. For each configuration we measure the wall-clock time of the decoding workload, normalize it to the vanilla $\Delta=32$ setting, and report the resulting relative throughput in Table~\ref{tab:gpu4perf}. The gains steadily increase as we parallelize more layers (e.g., $\Delta=23$ reaches $1.46\times$), illustrating the benefit of halving the number of synchronization steps per block; however, these aggressive settings correspond to the larger accuracy drops discussed in the main text, so moderate $\Delta$ values offer a better accuracy–speed trade-off.

\begin{table*}[h!]
\centering
\caption{\label{tab:gpu4perf}Relative throughput of 4-GPU Layer Parallelism on Llama 2 7B measured on $4\times$ A100 80\,GB PCIe GPUs. $\Delta$ denotes the number of layers replaced by LP pairs, with $\Delta=0$ serving as the baseline ($x1.00$).}
\begin{tabular}{lc}
\toprule
$\Delta$ & Rel. Speed \\
\midrule
0 & x1.00 \\
10 & x1.24 \\
14 & x1.34 \\
18 & x1.46 \\
\bottomrule
\end{tabular}
\end{table*}

\newpage
\section{Acceleration source}

\begin{figure}[htbp]
    \centering
    \begin{subfigure}[b]{0.8\textwidth}
        \centering
        \includegraphics[width=\textwidth]{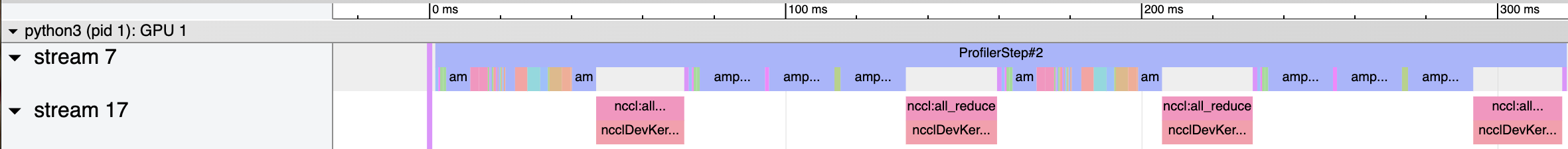}
        \caption{Flame chart of running two standard Tensor-Parallel LLama decoder layers.}
        \label{fig:tpprofile}
    \end{subfigure}
    
    \vskip\baselineskip 

    \begin{subfigure}[b]{0.8\textwidth}
        \centering
        \includegraphics[width=\textwidth]{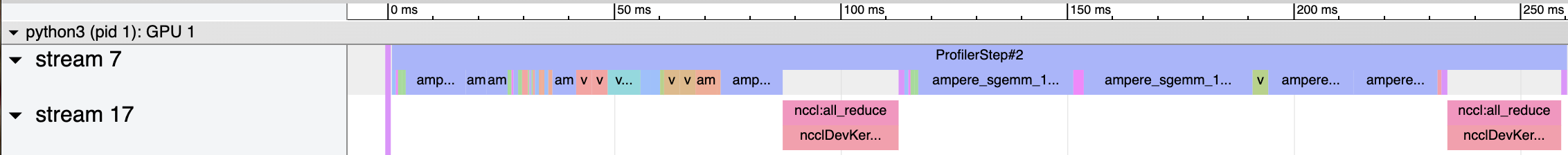}
        \caption{Flame chart of running two Llama 3 decoder layers with our Layer Parallelism approach.}
        \label{fig:tpparallelprofile}
    \end{subfigure}
    
    \caption{Comparison of Flame Graphs when running two consecutive Llama 3.2 3B decoder layers with vanilla tensor parallelism (Fig.~\ref{fig:tpprofile}), and our Layer Parallelism approach (Fig.~\ref{fig:tpparallelprofile}). Note that the time axis scale is different between both graphs. These results were obtained on a workstation using x2 RTX 4090s.}
    \label{fig:profiling}
\end{figure}

Figure~\ref{fig:profiling} illustrates flame graphs comparing two consecutive Llama 3.2 3B decoder layers using vanilla tensor parallelism and our LP approach. The profiling data summarized in Table~\ref{tab:profile-results} reveals that the primary source of acceleration in our LP method stems from reducing the total number of \texttt{all-reduce} synchronization operations across GPUs. Specifically, the vanilla tensor parallel approach performs synchronization at every decoder layer, resulting in higher cumulative synchronization overhead due to the fixed latency costs. Size-independent latency is dominated by GPU kernel-launch overhead, per-hop interconnect round-trip latency and host/stream synchronization imposed by the calling context. In contrast, our Layer Parallelism implementation runs pairs of layers simultaneously, effectively halving the number of synchronization points. This reduction in synchronization leads to a significant drop in synchronization time from 100.8ms to 50.7ms, directly contributing to the observed improvement in inference speed.

Additionally, Layer Parallelism enables fusion of certain computation kernels—particularly attention and MLP operations across parallelized layers—which further marginally reduces the computation time from 217ms to 208.7ms. Although these computational gains are modest compared to the savings achieved through fewer synchronization operations, kernel fusion further optimizes hardware utilization and enhances overall throughput.

\begin{table}[htbp]
\caption{Profiling results comparing vanilla Tensor Parallel and Layer Parallel implementations on two consecutive Llama 3.2 3B decoder layers.}
\centering
\begin{tabular}{lccc}
\toprule
\textbf{Approach} & \textbf{Total Time (ms)} & \textbf{Sync Time (ms)} & \textbf{Computation Time (ms)} \\
\midrule
Tensor Parallel & 317.8 & 100.8 & 217.0 \\
Layer Parallel (Ours) & 259.4 (x1.23) & 50.7 (x1.99) & 208.7 (x1.04) \\
\bottomrule
\end{tabular}
\label{tab:profile-results}
\end{table}

\end{document}